\documentclass{article} 
\usepackage{iclr2021_conference,times}

\usepackage{amsmath,amsfonts,bm}

\def\eqref#1{equation~\ref{#1}}

\def\1{\bm{1}}

\DeclareMathAlphabet{\mathsfit}{\encodingdefault}{\sfdefault}{m}{sl}
\SetMathAlphabet{\mathsfit}{bold}{\encodingdefault}{\sfdefault}{bx}{n}

\usepackage{hyperref}
\usepackage{url}

\usepackage[utf8]{inputenc} 
\usepackage[T1]{fontenc}    
\usepackage{hyperref}       
\usepackage{url}            
\usepackage{booktabs}       
\usepackage{amsfonts}       
\usepackage{nicefrac}       
\usepackage{microtype}      
\usepackage{graphicx}
\usepackage{xcolor}
\usepackage{subcaption}

\usepackage{amsmath}
\usepackage{amssymb}
\usepackage{wrapfig}
\usepackage{listings}

\let\oldparagraph=\paragraph
\renewcommand\paragraph[1]{\oldparagraph{#1.}}

\newcommand{\ignore}[1]{}

\newcommand{\x}{\mathbf{x}}
\newcommand{\y}{\mathbf{y}}
\newcommand{\m}{\mathbf{m}}
\newcommand{\constraint}{\mathcal{C}}

\title{Learned Hardware/Software Co-Design \\ of Neural Accelerators}
\begin{document}

\author{Zhan Shi, Chirag Sakhuja \\ 
The University of Texas at Austin \\
\texttt{\{zshi17, chirag.sakhuja\}@utexas.edu} \\
\And 
Milad Hashemi, Kevin Swersky \\ 
Google Research \\
\texttt{\{miladh, kswersky\}@google.edu} \\
\And
Calvin Lin \\
The University of Texas at Austin \\
\texttt{lin@cs.utexas.edu} \\
}

\maketitle

\begin{abstract}
The use of deep learning has grown at an exponential rate, giving rise
to numerous specialized hardware and software systems for deep learning.
Because the design space of deep learning software stacks and hardware
accelerators is diverse and vast, prior work considers software optimizations separately from hardware architectures, effectively reducing the search space.  Unfortunately, this bifurcated approach means that many profitable design points are never explored.  This paper instead casts the problem
as hardware/software co-design, with the goal of automatically identifying
desirable points in the joint design space.  The key to our solution is
a new constrained Bayesian optimization framework that avoids invalid
solutions by exploiting the highly constrained features of this design
space, which are semi-continuous/semi-discrete.  We evaluate our
optimization framework by applying it to a variety of neural models,
improving the energy-delay product by 18\% (ResNet) and 40\% (DQN) over
hand-tuned state-of-the-art systems, as well as demonstrating strong
results on other neural network architectures, such as MLPs and Transformers.

\end{abstract}

\section{Introduction}
\vspace{-.1in}

The compute requirements of deep learning are growing at a double
exponential rate~\citep{hern2020measuring}, with more powerful
models requiring exponentially more compute to train.  This growth has been
enabled by large systems of hardware accelerators, like GPUs and
TPUs~\citep{v100, jouppi2017datacenter}.  However, the continued scaling of
these systems is limited by issues of power density, cooling, and memory, so we need to improve computational efficiency.

Efficiency improvements can be sought at each layer of the deep learning
stack, from better learning algorithms~\citep{kingma2014adam}, to improved
neural network architectures~\citep{tan2019efficientnet}, to deep learning
compilers~\citep{chen2018learning}, to specialized DNN accelerators that
increase hardware efficiency~\citep{chen2014diannao, chen2016eyeriss}.
In this paper, we focus on the low-level software and hardware portions
of this stack, with the goal of automatically optimizing the $energy\times
delay$ product of executing a particular model on a hardware accelerator.
We consider two components from the deep learning stack:  the hardware
accelerator and the software compiler that maps a model onto that hardware.
This area is commonly referred to as hardware/software co-design, and since
it requires human expertise from multiple disciplines (software engineers,
compiler writers, hardware architects), it is typically driven by manual
heuristics or heuristic-based search~\citep{yang2020interstellar}.

We propose a different approach, recognizing that for a given DNN model,
this hardware/software co-design can be framed as a joint search of the
space of all of the valid mappings and hardware architectures that can
correctly execute the model.  We formally parameterize this space based
on prior work~\citep{parashar2019timeloop}, and we find that standard
optimization techniques, including off-the-shelf Bayesian optimization,
perform poorly because the design space is semi-discrete and the vast majority of the
points in the space are infeasible. Prior work~\citep{nardi2019practical} makes a similar observation, noting that (1) complex constraints such as hardware area and energy budget limit the feasible parameter values (i.e. small feasibility set), (2) some constraints are unknown until after a sample point has been evaluated (i.e. unknown feasibility).

\begin{figure}[t]
\centering
\includegraphics[width=1\textwidth]{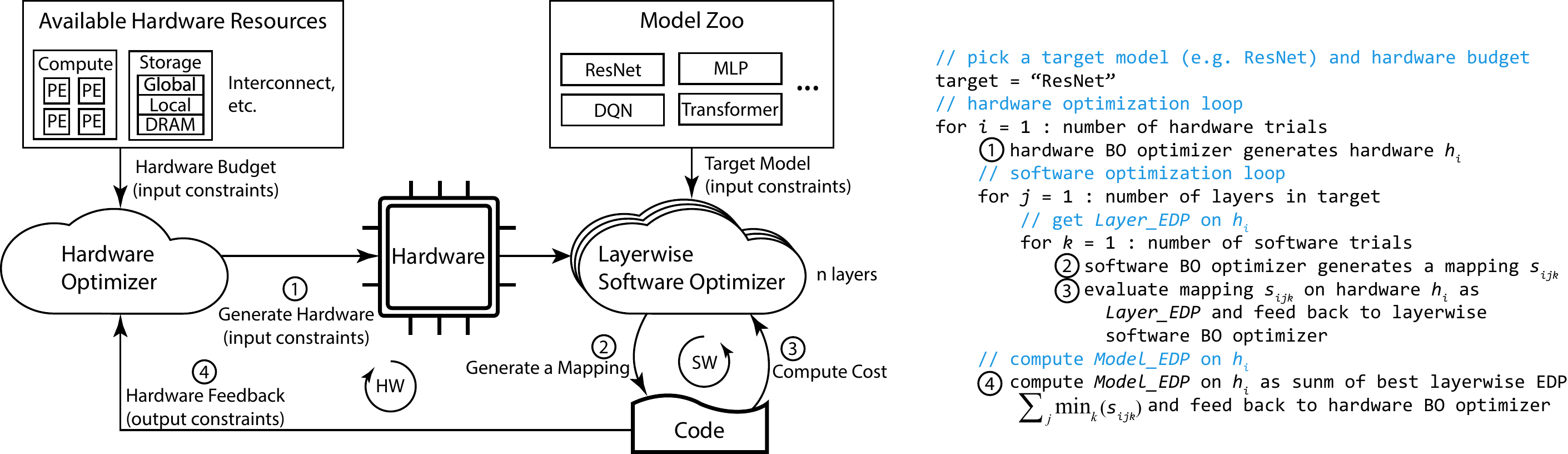}
\caption{Overview of BO-based nested search for hardware/software co-design.}
\vspace{-.2in}
\label{fig:model}
\end{figure}

Our solution casts the search as a bilevel optimization problem, as shown in Figure \ref{fig:model}.  The outer
loop optimizes over hardware architectures, while the inner loop optimizes
over software mappings for a given architecture.  Both of these are heavily
constrained black-box global optimization problems that require expensive
simulations to obtain performance estimates.  We therefore propose a nested,
constrained Bayesian optimization (BO) formulation that uses Bayesian
models of hardware and software performance to guide the search towards
promising regions of the design space.  Our approach is extensible to a
variety of different neural network architectures, and we make the code
publicly available.


We find that when compared against the state-of-the-art manually-designed
hardware accelerators that use heuristic software mappings, our BO-based
approach provides significant improvements in the speed and energy efficiency
of the resulting system, improving the energy-delay product (EDP) by 16.0\%
to 40.2\% on a series of neural networks.  The key to our solution is
our robust BO software optimizer, whose consistent efficiency allows our
approach to scale to this huge search space.

This paper makes the following contributions:
\vspace{-.1in}
\begin{itemize}
    \item We present the first system that automatically co-optimizes both the hardware and software portions of DNN accelerator design using a principled and systematic search algorithm.

    \item We present a constrained formulation of hardware and software design for BO, a challenging problem given the high ratio (90\%) of invalid hardware and software designs.

    \item We present a nested hardware/software formulation of BO that is extensible to other hardware accelerator designs.

    \item We provide model-specific hardware and state-of-the-art results on multiple
          models.
\end{itemize}

\vspace{-.2in}
\section{A Formal Representation of Software and Hardware}
\label{sec:hwsw_representation}
\vspace{-.1in}

Hardware/software co-design typically performed manually, but we believe that
this vast design space is best navigated by an intelligent search process.
To facilitate this automation, this section formally defines the hardware
and software design spaces.


\vspace{-.15in}
\subsection{Parameterizing the Design Space}
\vspace{-.1in}

\textit{Software design points} can be parameterized by the loop ordering, loop
tiling, and computational parallelism of the seven-level loop nest used to
compute a convolutional layer (see appendix), as has been
noted by recent work~\citep{parashar2019timeloop, yang2020interstellar}. These
software parameters are subject to hardware constraints, such as the quantity
and layout of processing elements (PEs) and the size of storage elements.

\textit{Hardware parameters} \ignore{are generally more specific to the
low-level resource and memory configurations or the layout of PEs.  These}
can be broken down into a two broad categories:

\textit{Resource configurations} represent the
physical aspects of hardware, such as buffer sizes, tile sizes, and the
cluster size of global buffers, as well as the layouts of the PE array
and of the global buffer.

\textit{Dataflow configurations} represent the usage
of the PE array that are implemented in hardware, such as the blocking
factors and degree of parallelism at the PE level, which also determines
the communication patterns among PEs.

\begin{figure}[ht]
  \centering
  \includegraphics[width=\textwidth]{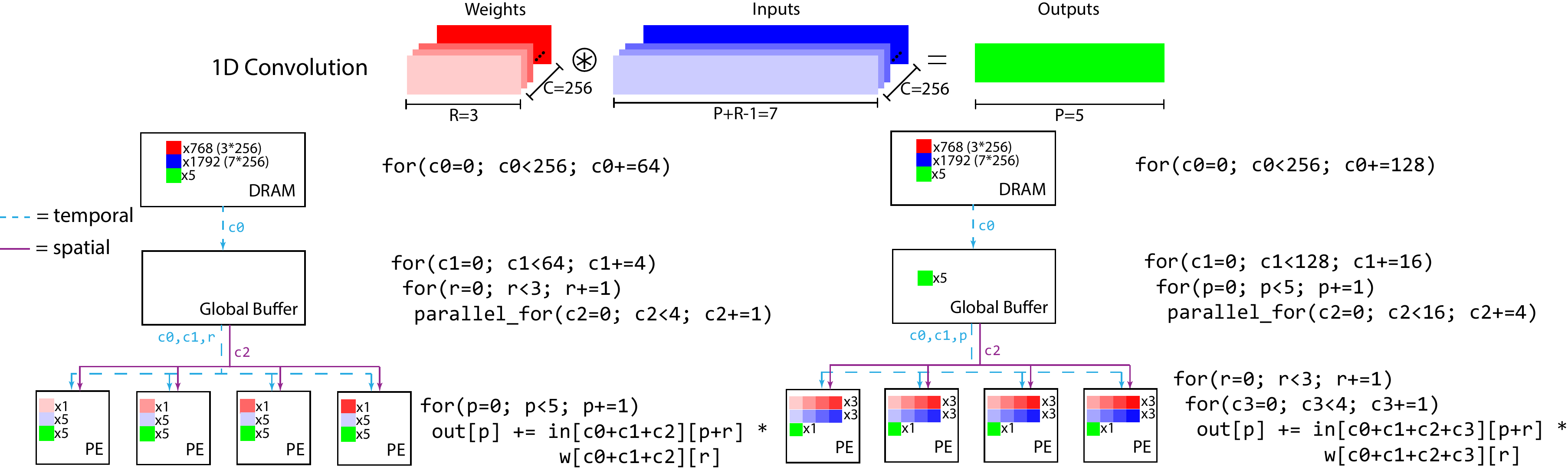}
  \caption{Two architectures computing a 1D convolution.}
  \label{fig:conv1d}
  \vspace{-.1in}
\end{figure}

Figure~\ref{fig:conv1d} shows two possible design points for a 1D
convolution.  Both design points tile and parallelize the channel (C)
dimension.  To the right of each component in the architecture is a set of
loops that specifies the control logic for the component, which can be broken
down into temporal streaming (\texttt{for} loops) and spatial distribution
(\texttt{parallel\_for} loops).  For example, in the architecture on the
left, the global buffer distributes across the PEs 1 weight from 4 separate
channels (\texttt{c2}), and the PEs perform all operations that the weight
participates in.  In this design, all data reuse is captured within the PE,
so the global buffer need not store anything.  By contrast, the architecture
on the right distributes a single output element across the PEs to compute
partial sums, which are stored in the global buffer across iterations.
Both these design points consist of the same architectural components,
but the dataflows vary, imposing different constraints on the software.
The appendix shows the details of the parameterization of a more practical
2D convolution.

\vspace{-.15in}

\vspace{-.1in}
\subsection{Constraints in the Design Space}
\vspace{-.1in}

There are several reasons why the vast space of hardware and software
parameters is filled with impractical or invalid design points.
First, hardware designs are fundamentally constrained by area (the total
amount of compute and storage resources) and factors such as available
memory bandwidth.  Second, the design cost and latency of additional area
grow super-linearly~\citep{shao2019simba}, which leads to many impractical
design points.

Software constraints are generally governed by feasibility instead of
practicality and predominantly depend on the hardware configuration and the
specific neural network workload.  For a specific hardware accelerator, there
is a limited number of available resources, so the software optimization
problem can be viewed as a search for the most efficient use of hardware
PEs and buffers.  For example, the loop blocking optimization factors a
neural network across multiple hardware storage buffers---and the feasible
factorizations are constrained by the size of the hardware buffers.


\vspace{-.175in}
\section{Bayesian Optimization}
\label{sec:bayesopt}
\vspace{-.125in}
\subsection{Overview}
\vspace{-.1in}

Bayesian optimization~\citep{jones1998efficient, brochu2010tutorial,
shahriari2015taking} is an effective approach for the optimization of
expensive, possibly noisy black-box functions.  BO has been used to 
optimize hyperparameters~\citep{snoek2012practical}, configure
algorithms~\citep{hutter2011sequential}, optimize A/B experiments
\citep{letham2019constrained}, and more.  For our problem, we have a
parameterized representation and access to a simulator. Since one of our
main concerns is sample efficiency, Bayesian optimization is particularly
suitable.

The actual cost of evaluation depends on the experimental infrastructure, but in general, it is much more expensive to evaluate a hardware design choice than to evaluate software optimizations, because hardware design can take hours (to produce a hardware simulator or an FPGA) to days or even months (to produce an ASIC).



Bayesian optimization has two major components: (1) a surrogate model
provides a Bayesian posterior probability distribution that predicts
potential values of the objective function.  (2) an acquisition function
uses the model to identify the next point to evaluate.


\vspace{-.15in}
\subsection{Gaussian processes}
\vspace{-.1in}

A common surrogate model is a Gaussian process (GP)~\citep{gpml2006book} due to its simplicity and flexibility. A GP is prior distribution over the space of functions comprised of a mean function $m(\x)$ and a covariance, or kernel function $k(\x, \x')$. Suppose we are given a dataset of $N$ input/output pairs over a bounded domain $\Omega$ with $D$ input dimensions and scalar outputs. For brevity, we write this as $(X,\y)$, where $X \in \Omega^{N \times D}$ and $\y \in \mathbb{R}^N$. The posterior predictive distribution over function values $f$ for a new input $\x$ is given by $P(f \mid \x, X, \y) = \mathcal{N}(\mu(\x), \sigma^2(\x))$, where
\begin{align*}
    \mu(\x) &= K_{\x X} K_{XX}^{-1}(\y - \m_X) + m(\x), \\
    \sigma^2(\x) &= k(\x, \x) - K_{\x X}K_{XX}^{-1}K_{\x X}^\top.
\end{align*}
Where $K_{XX}$ is a matrix formed by evaluating the kernel on $X$, $K_{\x X}$ is the vector of kernel evaluations between $\x$ and $X$, and $\m_X$ is the vector of mean function evaluations on the input dataset.

A common choice for the kernel is squared exponential. Given two input vectors $\x_i$ and $\x_j$, this is defined as $k(\x_i, \x_j) = \alpha^2 \exp\left (-\frac{\|\x_i - \x_j\|^2}{\boldsymbol{\ell}^2}\right )$. $\alpha$ and $\boldsymbol{\ell}$ are kernel hyperparameters.

Another kernel that we find particularly useful is a linear kernel on top of explicit features. Given a feature mapping $\phi(\x): \mathbb{R}^D \rightarrow \mathbb{R}^K$, the linear kernel can be written as $k(\x_i, \x_j) = \phi(\x_i)^\top \phi(\x_j)$. When we have strong prior information about the relevant feature interactions that govern the black-box function, this kernel allows us to encode these interactions directly and produces a more sample-efficient posterior estimate.

In cases where the observations from the black-box function are noisy, we can add a noise kernel $K_\mathrm{noise} = \tau^2 \mathrm{I}$ to $K_{XX}$, where $\tau^2$ is a hyperparameter. This implies a Gaussian observation likelihood.

Following common practice, we use the constant mean $m(\x) = c \quad \forall \  \x$. All kernel and mean hyperparameters are learned by maximizing the marginal likelihood of the GP on the current dataset.

\vspace{-.1in}
\subsection{Acquisition functions}
\vspace{-.1in}

A critical component in the BO framework is the choice of acquisition function $a(\cdot)$ that assigns each design point a value that represents the utility of testing this point. Two commonly used acquisition functions are expected improvement (EI) and lower confidence bound (LCB).

EI computes the amount we expect to improve upon the current best observed objective value $y_* \equiv \mathrm{max}\{y_i\}_{i=1}^{N}$ by evaluating a design point $\x$. Formally, it can be written as
\begin{align*}
    a_{\mathrm{EI}}(\x) &= \int_{-\infty}^{\infty}\mathrm{max}(y_* - f, 0) P(f \mid \x, X, \y)\mathrm{d}f.
\end{align*}

\vspace{-.05in}
where $f$ is the latent function from the surrogate model, and $y_*$ is the best value observed.


LCB~\citep{srinivas2009gaussian} provides an explicit tradeoff between the predictive mean and variance and is defined as
\vspace{-.05in}
\begin{align*}
    a_\mathrm{LCB}(\x) &= \mu(\x) + \lambda \sigma(\x).
\end{align*}
Where $\lambda$ represents a tradeoff parameter. A small $\lambda$ promotes greater exploitation, and a large $\lambda$ promotes greater exploration. We found $\lambda=1$ to work well in our experiments. Beyond these, there are many other possible acquisition functions that could be used in future exploration~\citep{thompson1933likelihood, hennig2012entropy, hernandez2014predictive, frazier2009knowledge}.
\vspace{-.15in}
\subsection{Constraints}
\vspace{-.1in}

In our problem, the vast majority of the design space will produce invalid solutions.
When the constraints are a known function of the input features, we can directly account for them as input constraints. Otherwise, we must run the simulation and treat invalid points using an output constraint. Here, we will describe these constraint types, and how they are incorporated into BO.

\emph{Input constraints} are explicit constraints that are used when optimizing the acquisition function. They directly prevent the search from suggesting points that will violate the constraints. As some constraints are non-linear, this optimization is itself very challenging, as it is a global optimization problem with non-convex constraints. In the unconstrained case, maximizing the acquisition function often takes a hybrid approach: generating a random initial set of points and refining them by gradient ascent. Maintaining feasibility with non-convex constraints is far more challenging, however.

We therefore optimize the acquisition function in a simple way by performing rejection sampling on the design space: we randomly sample parameters until we obtain 150 feasible points, and then choose the one the maximizes the acquisition function. On average the sampling takes 22K random samples to get a pool of 150 feasible points. We have found that practically this is a simple yet effective strategy for our problems, we leave more involved optimization schemes for future work.

\emph{Output constraints} are used when we do not know the form of the constraint a-priori and must run the simulator to test feasibility. This is also referred to as an ``unknown'' constraint, and BO has been adapted to incorporate a constraint model in addition to the regression model~\citep{gelbart2014bayesian}. These simultaneously learn about the constraint boundaries while modeling the objective.

Let $\constraint(\x)$ denote the event that $\x$ satisfies constraint $\constraint$. Constrained BO uses a Bayesian classifier to model $P(\constraint(\x))$. It is relatively straightforward to adapt a GP regressor to 
classification~\citep{gpml2006book}.

Under a Bayesian classifier, the acquisition function $a(\x)$ is modified to account for the probability that the constraint is satisfied, with 0 utility if it is not satisfied.
\begin{align*}
    \bar{a}(\x) &= \mathbb{E}[a(\x)\mathrm{I}[\constraint(\x)]] = P(\constraint(\x))a(\x).
\end{align*}
Where $\mathrm{I}[\constraint(\x)]$ is the indicator function that evaluates to 1 if the constraint is satisfied and 0 otherwise.
We therefore maintain two models: one regression model to capture the objective and one classifier to model the constraint in order to avoid evaluations in infeasible regions.

\vspace{-.1in}
\section{Bayesian Optimization for Hardware/Software Co-design}
\vspace{-.1in}

\subsection{Overview of Nested Hardware/Software Optimization}
\vspace{-.1in}
Provided the constraints discussed in Section \ref{sec:hwsw_representation} and the BO formulation from Section \ref{sec:bayesopt}, we propose a nested approach for co-optimizing hardware/software parameters. The overall approach is outlined in Figure~\ref{fig:model}. The goal is to find the optimal hardware parameters for a neural model and the optimal set of software parameters for each layer in the neural model. Since software constraints depend on a feasible hardware design, we first propose the hardware parameters, then for that hardware co-optimize the software mapping. 

Specifically, let $\x_h$ and $\x_s$ denote the set of hardware and software parameters in the parameter space to be optimized. In the nested search process, we first use the hardware optimizer to generate a design of hardware. In particular, we perform the hardware search in the space of possible hardware $\mathcal{S}_h$ to optimize all hardware parameters, where the objective is to minimize $f(\x_h \mid \textrm{NN})$ which we define as the energy-delay product (EDP) of running the neural network (NN) model on the given hardware, assuming the optimal software mapping for each individual layer. This step produces a hardware specification and can be formalized as $\textrm{{argmin}}_{h\in\mathcal{S}_h} f(\x_h \mid \textrm{NN})$.

For the chosen hardware design, our framework performs the software search for each individual neural layer in its constrained software mapping space $\mathcal{S}_s|h, \textrm{NN}_j$ to optimize the mapping parameters, where $\textrm{NN}_j$ denotes the $j$th layer in the neural network model, and the objective becomes $f(\x_s \mid \x_h, \textrm{NN}_j)$, which is defined as the EDP of running the layer $j$ on the fixed hardware. This step produces a design point that represents the best set of software mappings for each layers on the given hardware structure, and can be formalized as $\textrm{{argmin}}_{s\in\mathcal{S}_s|h} f(\x_s \mid \x_h)$. The layerwise EDPs are then summed up as the EDP of the neural model, which is fed back to the hardware optimizer to generate the next hardware setting.

The iterative search between hardware and software will repeat for a user-defined number of trials. In this work, we set 50 for hardware search and 250 for software search. The combination of hardware and software that achieves the best EDP during the optimization process becomes the final model-specific hardware structure and layer-specific software mappings. A random sample is used in the first iteration of both the hardware and software search. In our Bayesian optimization (BO) framework, we use separate BO models to search in the hardware and software space. We now describe their design considerations, particularly the choice of kernel and feature transformation.
\vspace{-.1in}
\subsection{BO for Optimizing Hardware Architectures}
\vspace{-.1in}
\paragraph{Kernel design} The main design choice for BO is the GP kernel to use. For the hardware search, we choose a linear kernel on top of feature transformations that represent the relationship between the different parameters. This feature transformation allows us to explicitly encode domain knowledge. For example, by comparing the layout parameters of the 2D PE array and global buffer we can obtain the ratio between these adjacent storage layers, which correlates to the maximal degree of parallel buffer accesses in each dimension. The details of the features are given Figure~\ref{fig:features} in the appendix. We also add a noise kernel to deal with noise in the hardware evaluation. This is because the software optimizer is not guaranteed to find the best software mapping for each layer. There is some randomness in the software search, and therefore independent runs of software optimization for a fixed hardware design may yield different results.

\vspace{-.1in}
\paragraph{Constraints} There are both known and unknown constraints in the hardware search. The known constraints, such as the compute and storage budget, are treated as input constraints that reject invalid samples. The unknown constraints have to do with feasibility (if there exist valid software mappings of neural layers onto the hardware, and if the valid mappings can be sampled during the software optimization). Following Section~\ref{sec:bayesopt}, these constraints are treated as output constraints and are modeled by a GP with a squared exponential kernel. 
\vspace{-.1in}
\subsection{BO for Optimizing Software Mappings}
\vspace{-.1in}
\paragraph{Kernel design} Similar to hardware optimization, we use a linear kernel and transform the parameters to features that encode relational information. As the hardware is fixed in the search of software mappings, we are able to compute features such as buffer usage, which potentially help make the predictions more accurate. The evaluation of a mapping on a given hardware is deterministic in our infrastructure, thus there is no need for a noise kernel in the GPs.

\vspace{-.1in}
\paragraph{Constraints} As both the hardware and neural model are known during software optimization, all constraints are known and are treated as input constraints that automatically reject invalid samples.

\vspace{-.15in}
\section{Evaluation}
\vspace{-.1in}
\subsection{Methodology}
\vspace{-.1in}
\paragraph{Infrastructure} We conduct our evaluation on
Timeloop~\citep{parashar2019timeloop}, which is an open-source infrastructure
for evaluating the hardware design and software optimization of DNN
accelerators.  Timeloop represents the key architecture attributes of DNN
accelerators that realize a broad space of hardware structure and topology,
which generate an accurate projection of performance and energy efficiency
for DNN workloads.  In the evaluation, Timeloop takes two inputs: 1) the
hardware configuration, which consists of the hardware-related parameters,
and 2) the software mapping, which consists of the software parameters
that describe the mapping.  As most accelerators are designed for neural
network inference, we limit the use case to inference in this work and
leave training for future work.

\vspace{-.1in}
\paragraph{Workloads}  To show that our solution automatically produces efficient hardware for a variety of neural networks,
we use our BO framework to optimize critical layers from CNNs
(ResNet~\citep{he2016deep} and DQN~\citep{mnih2013playing}), as well as an
MLP and Transformer~\citep{vaswani2017attention}.

\vspace{-.1in}
\paragraph{Experimental Setup}  We use Eyeriss~\citep{chen2016eyeriss},
a state-of-the-art DNN accelerator, as our baseline.  All workloads are
evaluated on the Eyeriss implementation with 168 PEs~\citep{chen2016eyeriss}
except for the Transformer model, which runs on the larger version of Eyeriss
with 256 PEs~\citep{parashar2019timeloop}.  In the software mapping search,
we use Eyeriss's hardware specifications and search for the best software
mapping for each neural layer.  In the hardware search, we perform the search
under the same compute and storage resource constraints as Eyeriss for each
neural model.~\footnote{This work focuses on model-specific hardware, but
hardware specialization provides larger benefits at a finer granularity,
i.e. if different layers can execute on customized hardware. We leave
this for future work.} 

\vspace{-.1in}
\paragraph{Metrics} Hardware accelerators are designed to achieve both speed
and energy efficiency, so we adopt the widely used energy-delay product
(EDP) as the objective.  Since EDP values can vary by an order of magnitude,
we normalize by dividing by the best (minimal) EDP value and take the
reciprocal for optimization curves.  For the hardware/software co-design, we
report the EDP improvements of each neural model, which is averaged across
all layers (see~Figure \ref{fig:cnn_layers} and \ref{fig:other_layers} in
the appendix).  For software mapping optimizations, we report the layer-wise
EDP improvements.

\vspace{-.1in}
\paragraph{Baselines} In hardware search, we compare against constrained random search that repeatedly takes the first random sample in the design space that satisfies the constraints.  In software search, we use constrained random search, TVM with XGBoost and TreeGRU~\citep{chen2018learning}, and out-of-the-box BO that optimizes in a continuous parameter space and rounds to the nearest valid parameters.

\vspace{-.1in}
\subsection{Software Mapping Optimization}
\begin{figure}[h]
\centering
\vspace{-.1in}
\begin{subfigure}{.24\textwidth}
  \centering
  \includegraphics[width=1\textwidth]{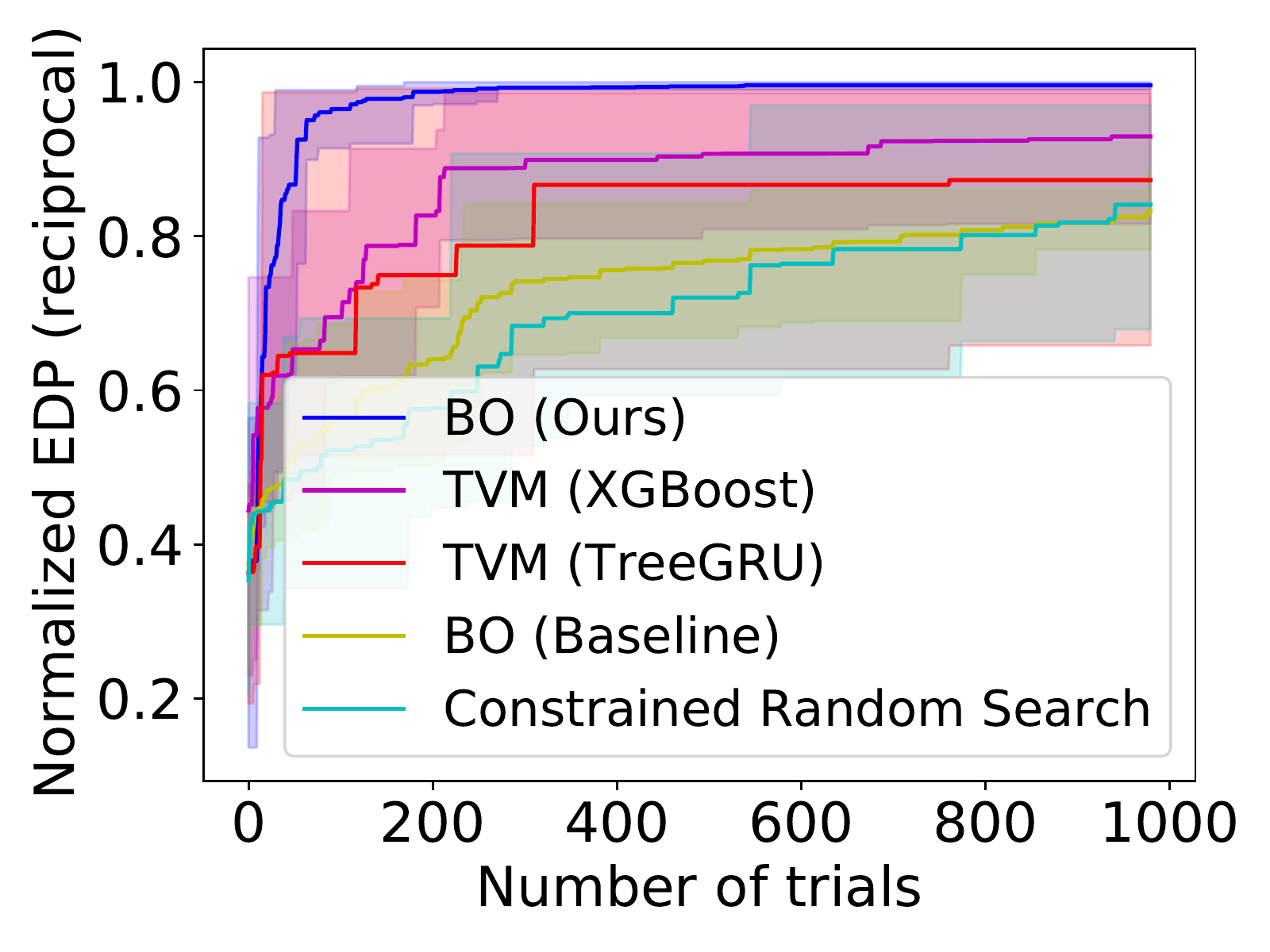}
  \caption{ResNet-K2}
\end{subfigure}
\begin{subfigure}{.24\textwidth}
  \centering
  \includegraphics[width=1\textwidth]{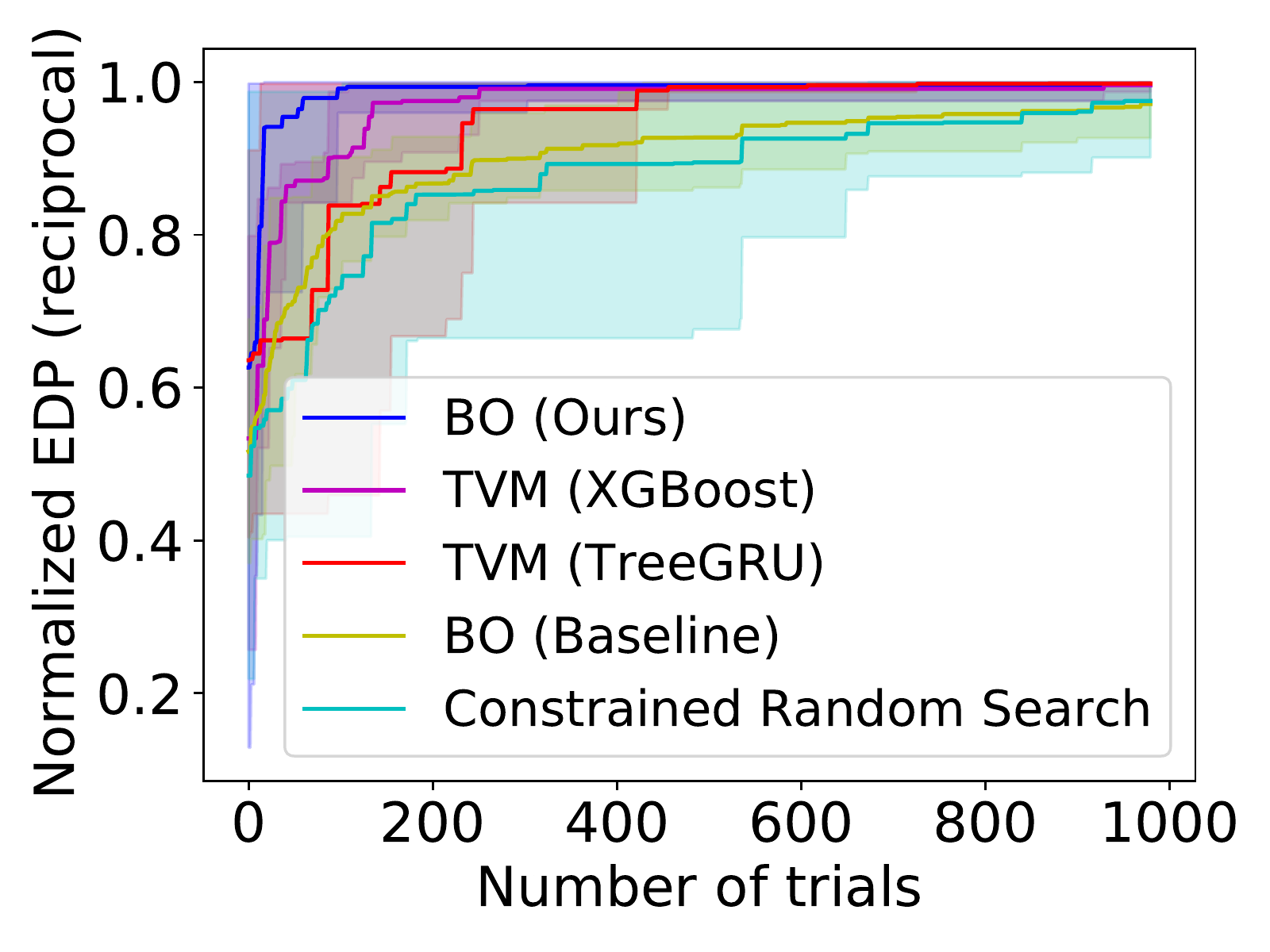}
  \caption{DQN-K2}
\end{subfigure}
\begin{subfigure}{.24\textwidth}
  \centering
  \includegraphics[width=1\textwidth]{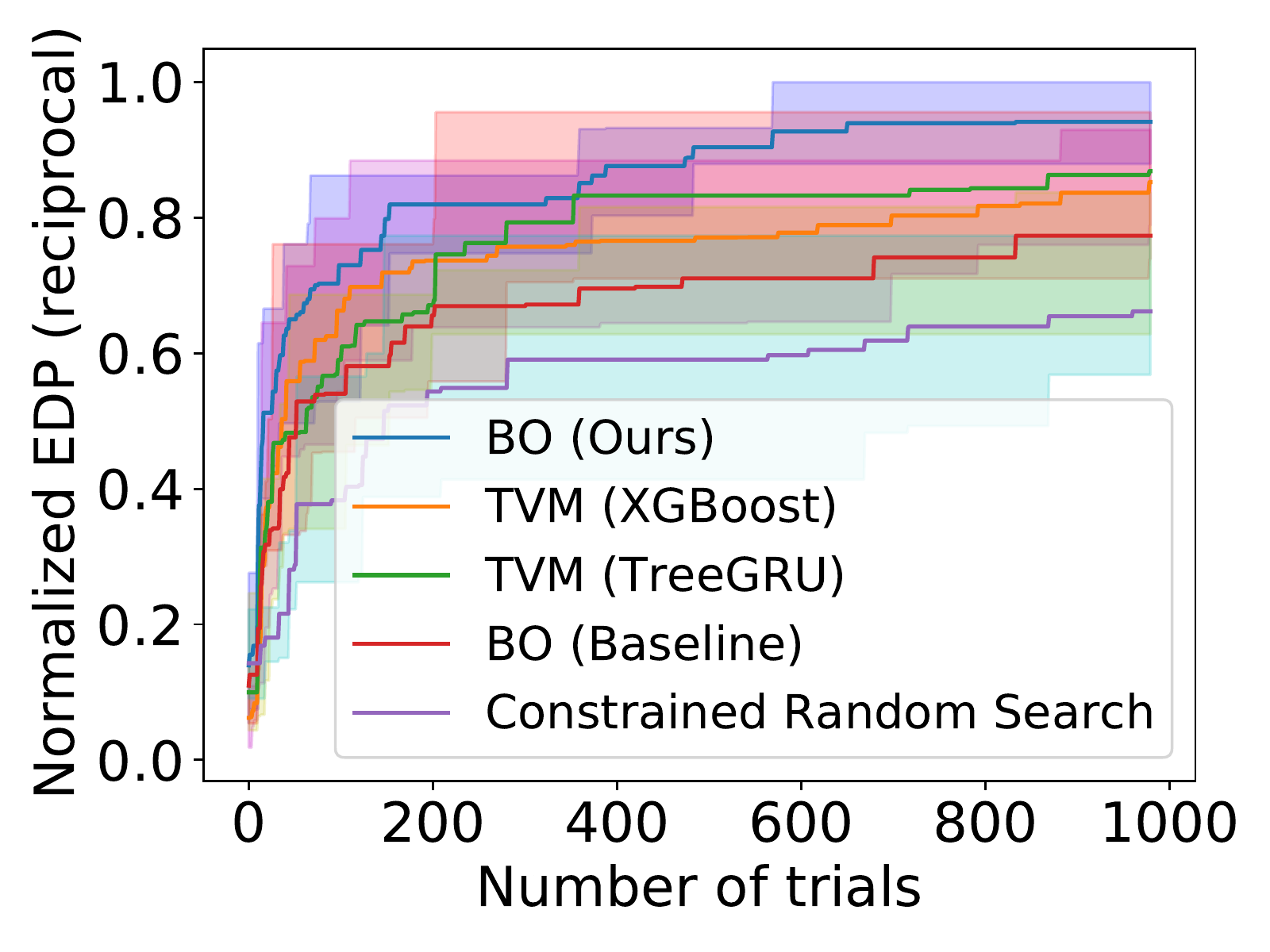}
  \caption{MLP-K2}
\end{subfigure}
\begin{subfigure}{.24\textwidth}
  \centering
  \includegraphics[width=1\textwidth]{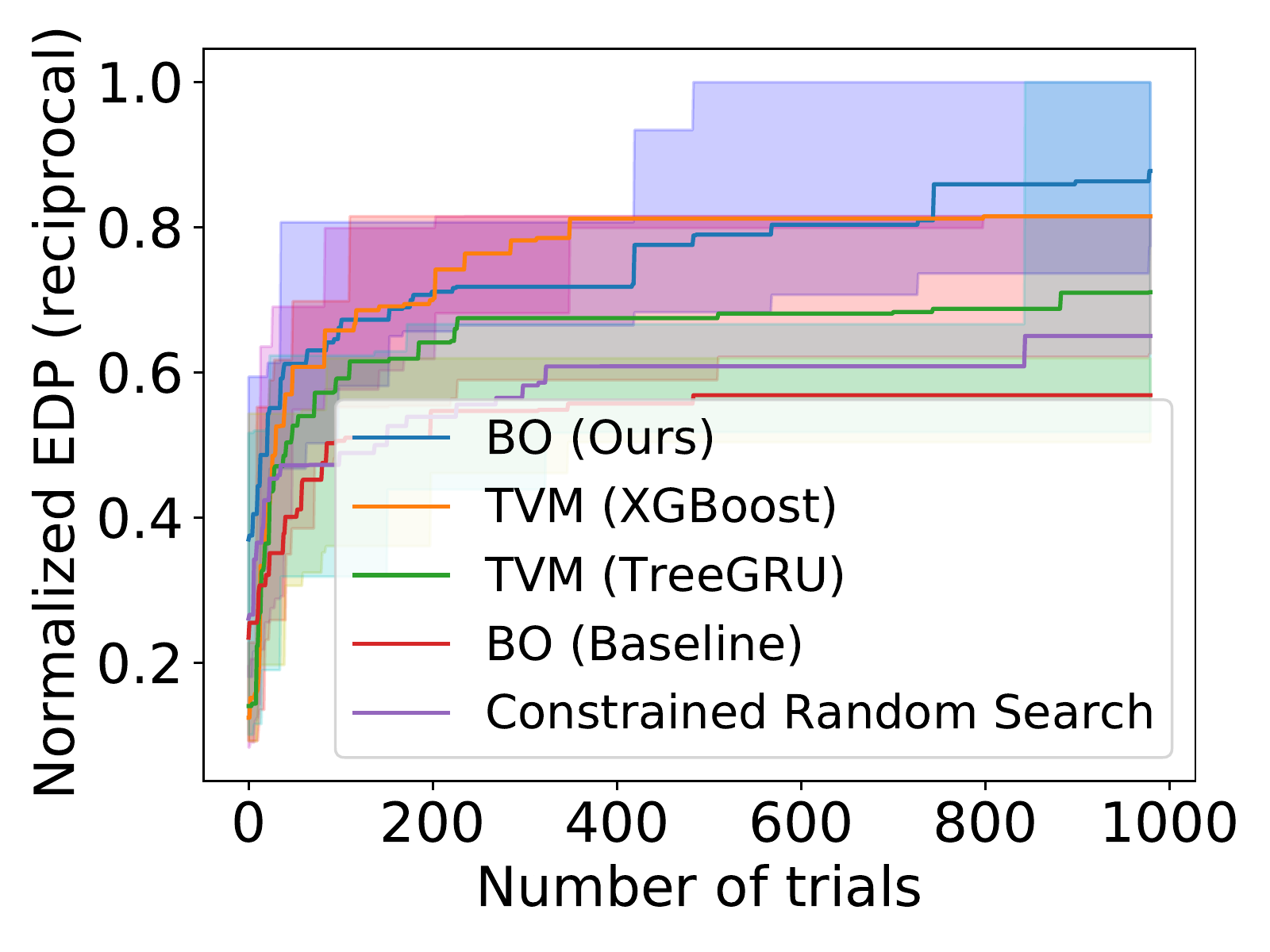}
  \caption{Transformer-K2}
\end{subfigure}
\caption{Software mapping optimization on layer 2 of ResNet, DQN, MLP, and Transformer. The y-axis shows the reciprocal of energy-delay product (EDP) (normalized against the best EDP value). Higher is better. Results for other layers can be found in the appendix. Best viewed in color.}
\vspace{-.2in}
\label{fig:sw_curves}
\end{figure}

\vspace{-.1in}
We show the results of software mapping optimization first, as the
capability of finding a good mapping is the base of evaluating a hardware
design.  Figure~\ref{fig:sw_curves} shows the improvements of BO over
our constrained random search formulation.  Our BO formulation outperforms random search, both variants of TVM as well as a standard BO formulation that optimizes discrete parameters using a relax-and-round approach.

\subsection{Hardware Configuration Optimization}
\begin{figure}[ht]
\centering
\begin{subfigure}{.24\textwidth}
  \centering
  \includegraphics[width=1\textwidth]{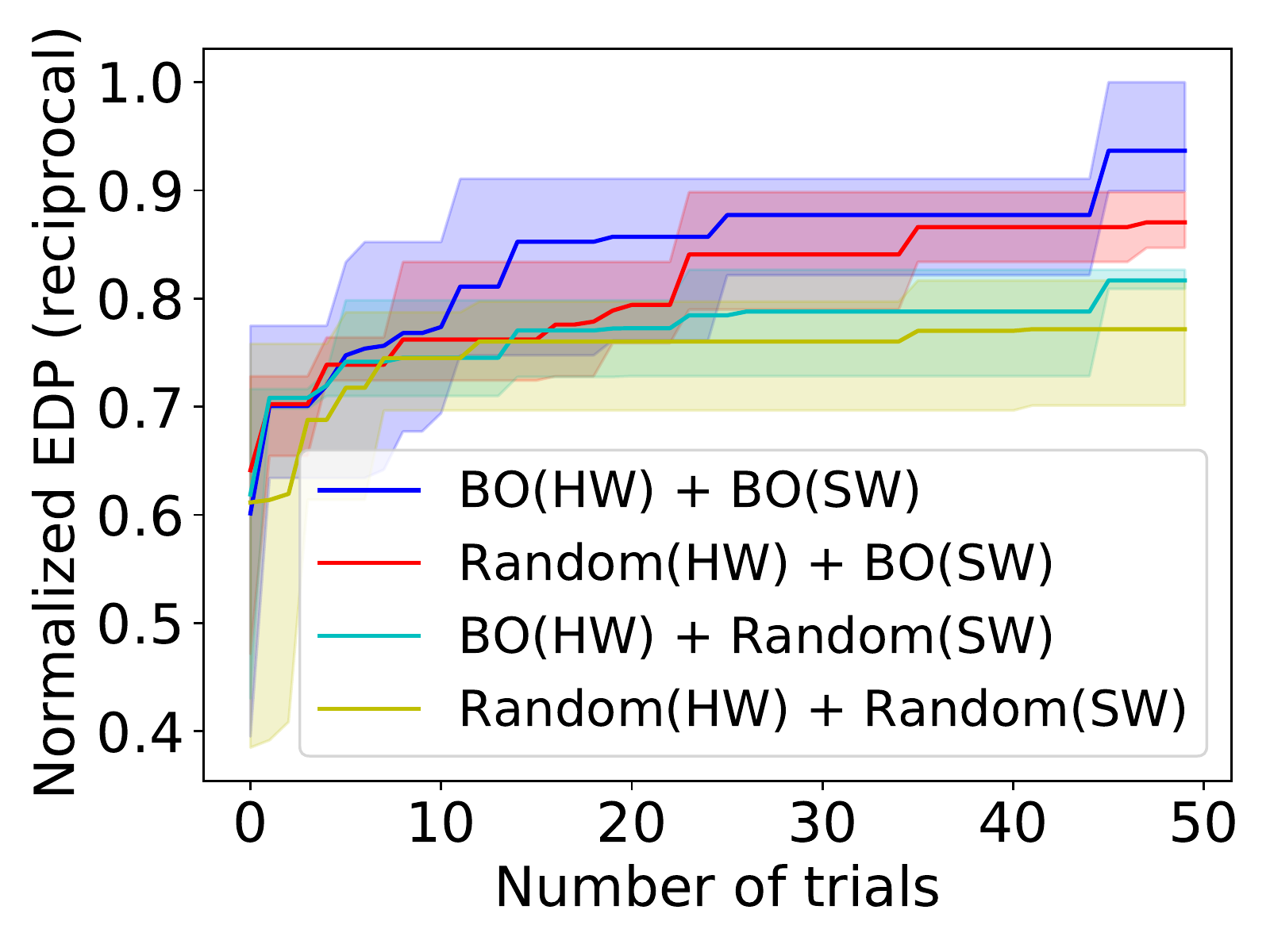}
  \caption{ResNet}
\end{subfigure}
\begin{subfigure}{.24\textwidth}
  \centering
  \includegraphics[width=1\textwidth]{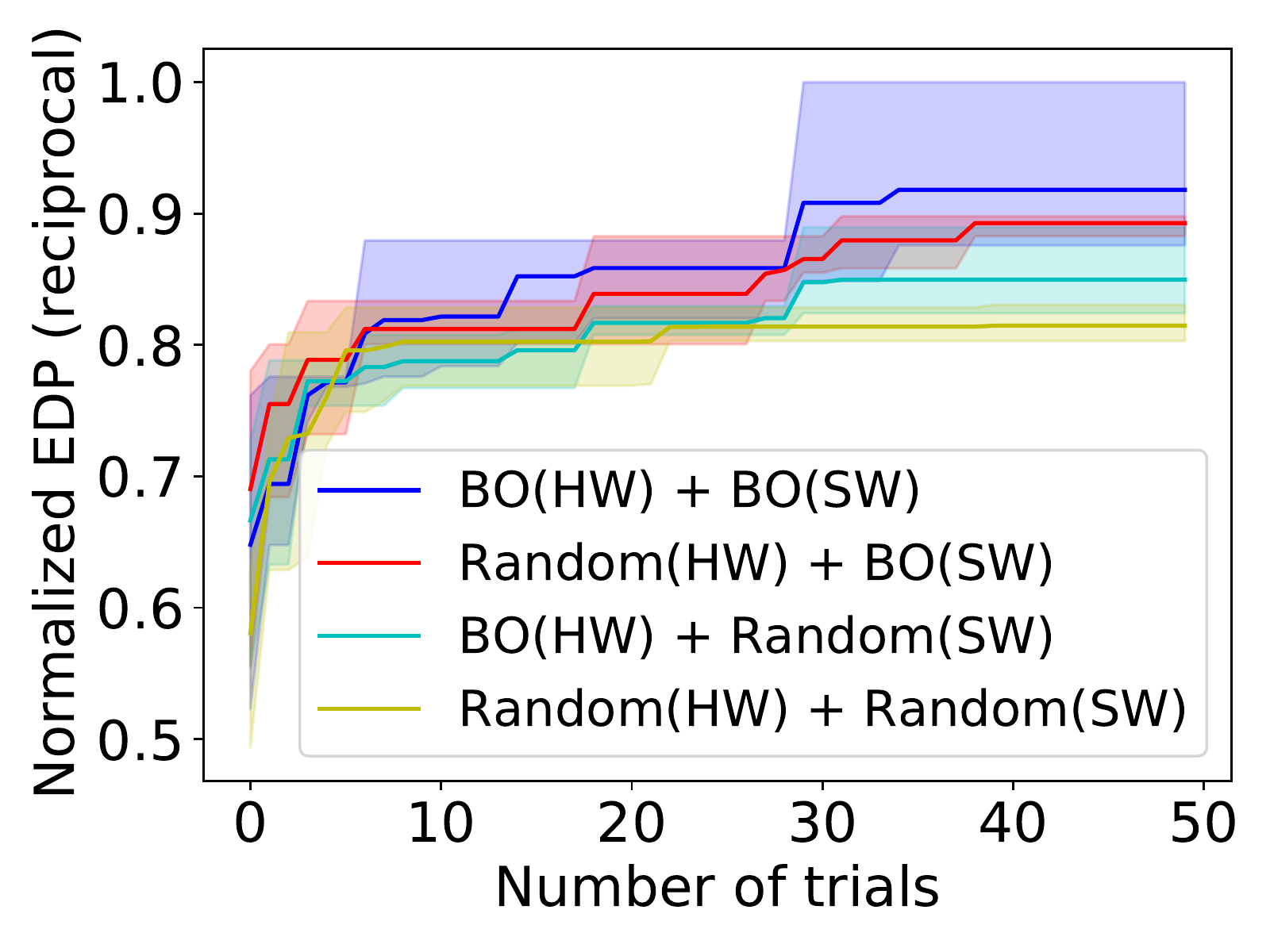}
  \caption{DQN}
\end{subfigure}
\begin{subfigure}{.24\textwidth}
  \centering
  \includegraphics[width=1\textwidth]{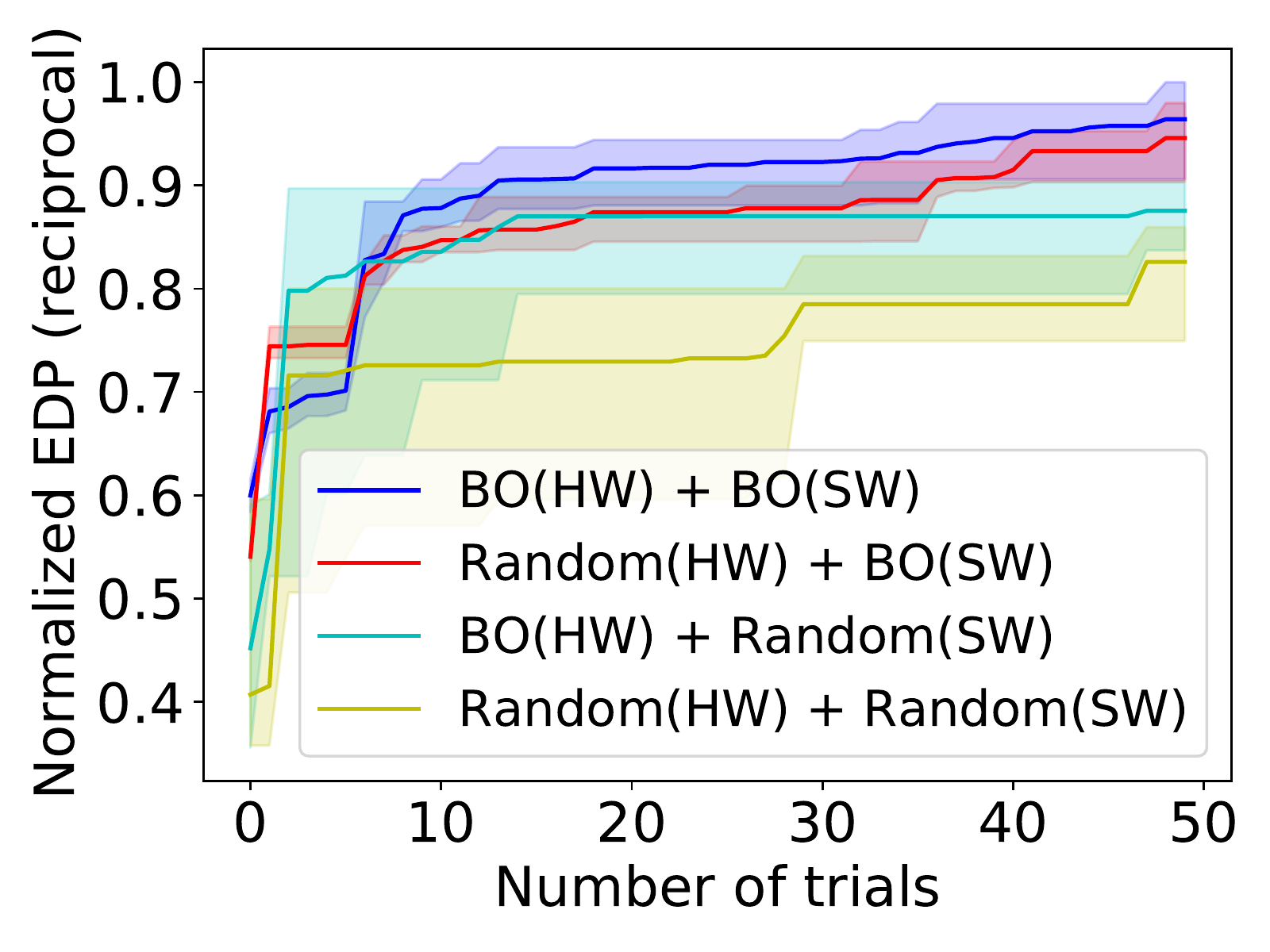}
  \caption{MLP}
\end{subfigure}
\begin{subfigure}{.24\textwidth}
  \centering
  \includegraphics[width=1\textwidth]{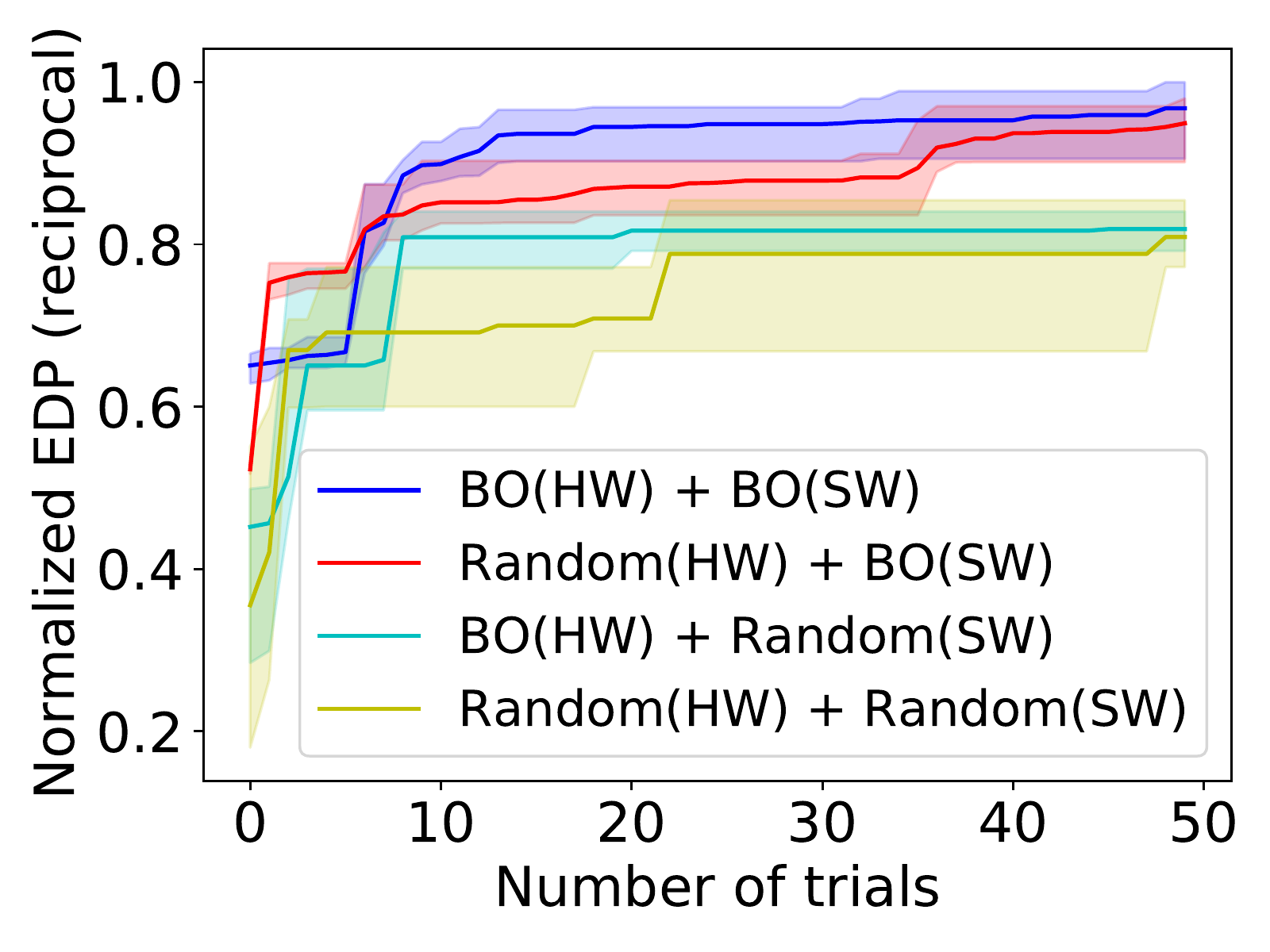}
  \caption{Transformer}
\end{subfigure}
\caption{Hardware/software co-optimization. The x-axis shows the number of trials for hardware search, and 250 attempts are made to find the optimal software mapping for each layer in the model on the hardware specification. Best viewed in color.}
\vspace{-.2in}
\label{fig:hw_curves}
\end{figure}

\vspace{-.1in}
Figure~\ref{fig:hw_curves} shows the optimization curves for hardware/software co-design. The comparison of hardware search algorithms shows that BO provides consistently better performance than the constrained random search, and the comparison of software search algorithms shows the importance of mapping optimization in the co-design process. As shown in Figure~\ref{fig:hw_compare}, we find that the designs searched by BO achieve significantly better EDP on all neural models compared to the state-of-the-art manually designed accelerator (18.3\%, 40.2\%, 21.8\% and 16.0\% for ResNet, DQN, MLP and Transformer respectively).

\vspace{-.05in}
\subsection{Ablation Studies}
\begin{figure}[ht]
\centering
\vspace{-.1in}
\begin{subfigure}{0.33\textwidth}
  \includegraphics[width=1\textwidth]{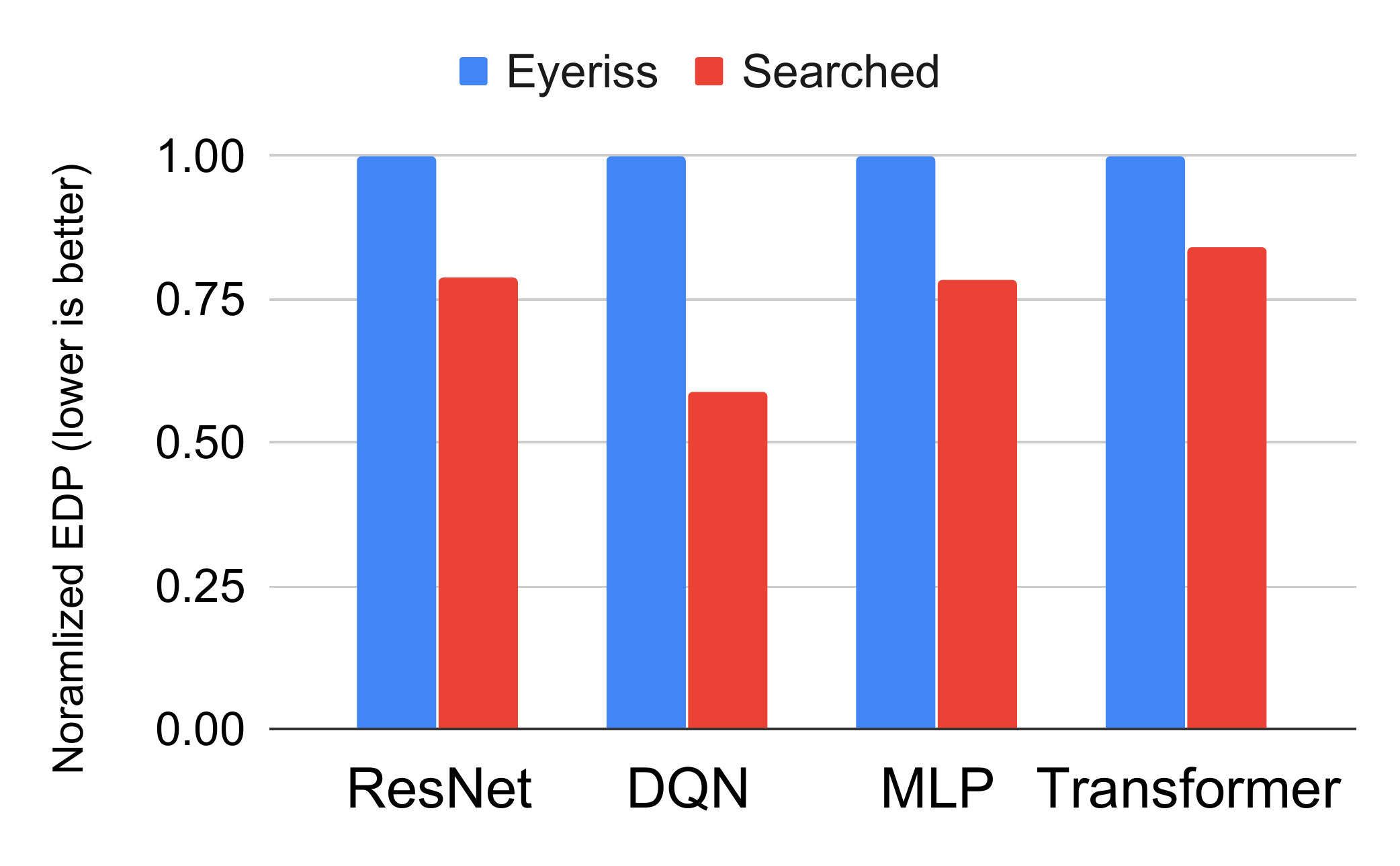}
  \caption{Comparison with Eyeriss.}
  \label{fig:hw_compare}
\end{subfigure}
\begin{subfigure}{0.3\textwidth}
  \centering
  \includegraphics[width=1\textwidth]{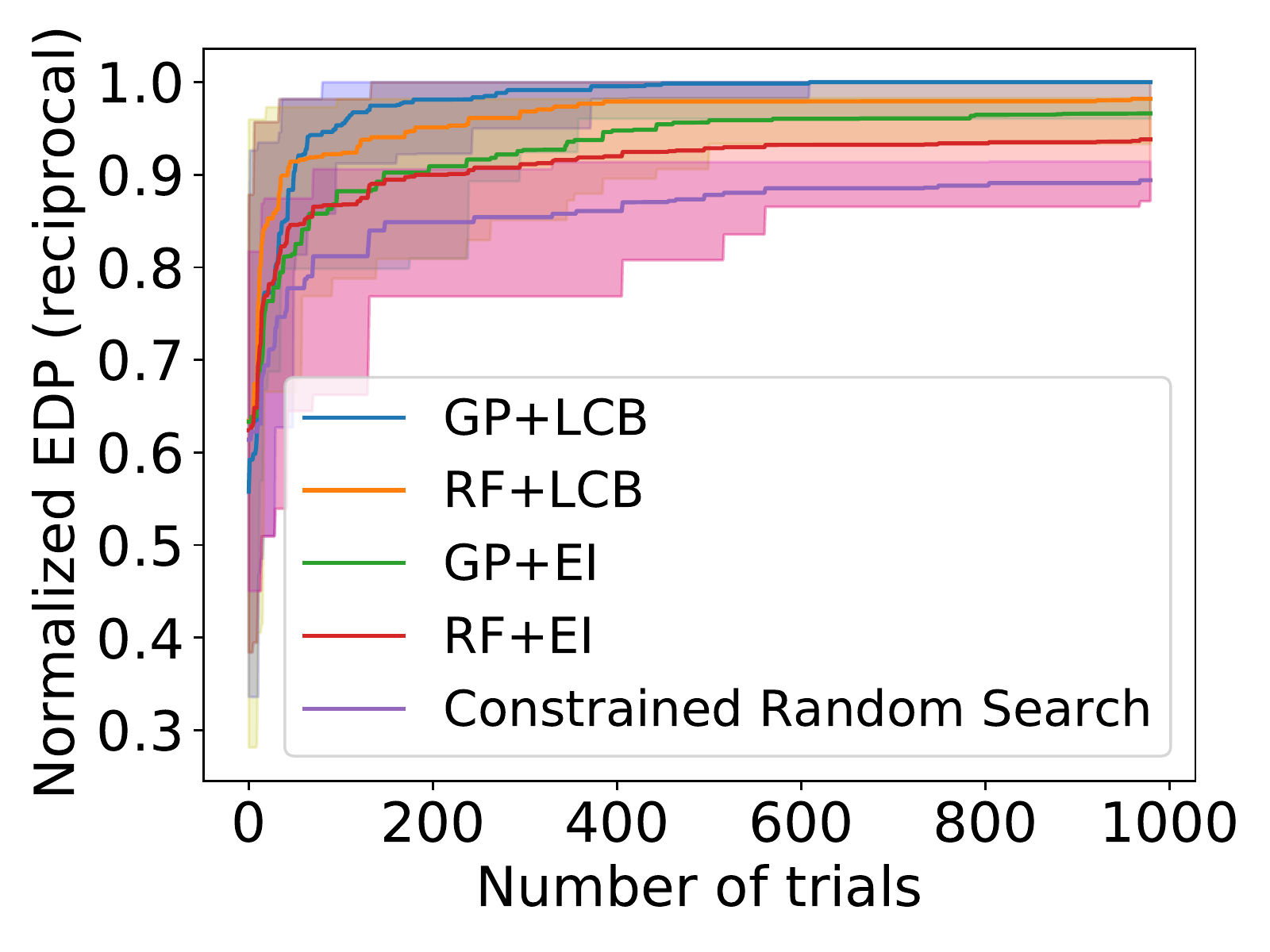}
  \caption{BO ablation}
  \label{fig:bo_ablation}
\end{subfigure}
\begin{subfigure}{.3\textwidth}
  \centering
  \includegraphics[width=1\textwidth]{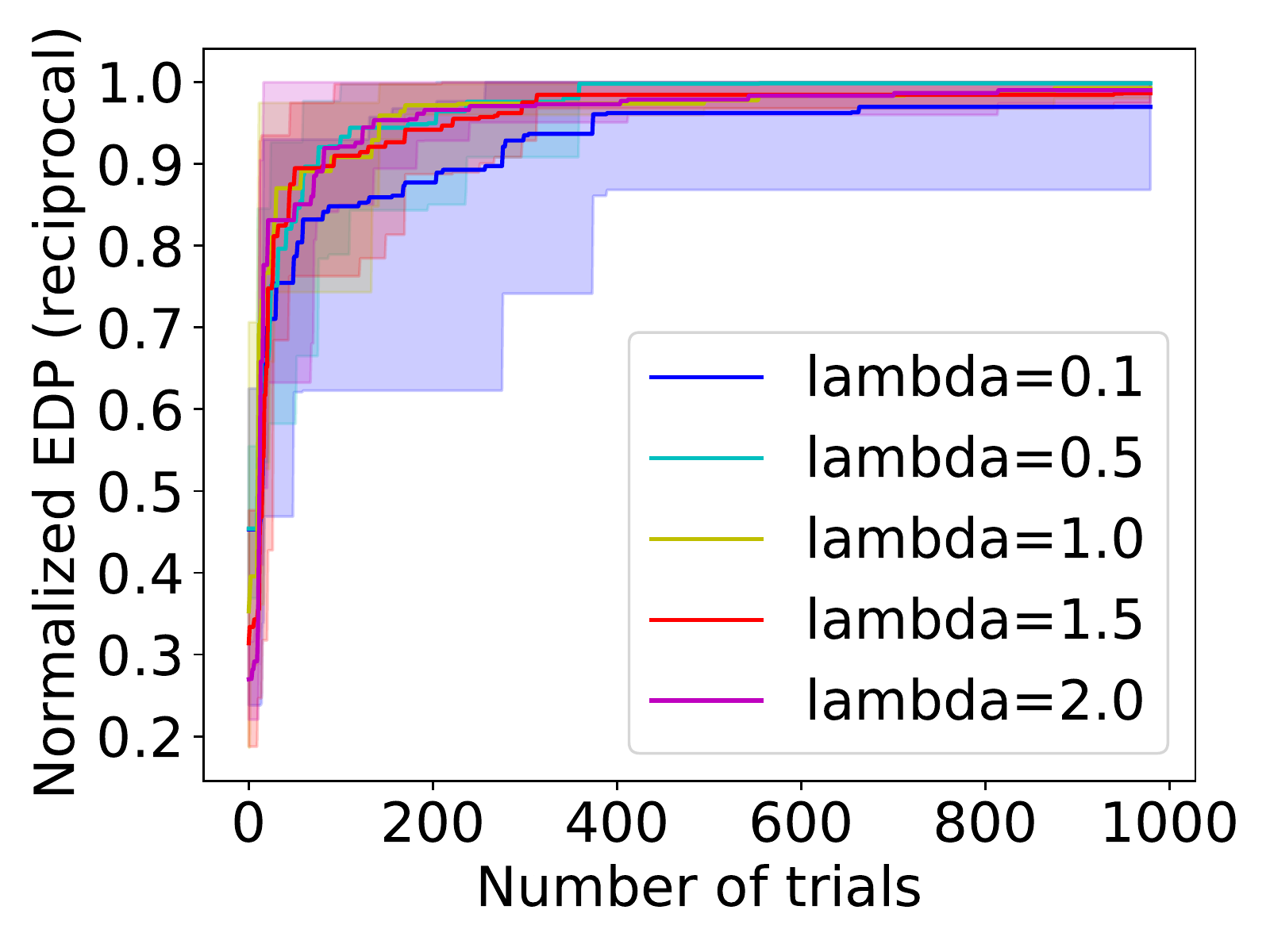}
  \caption{LCB $\lambda$ ablation}
  \label{fig:lcb_ablation}
\end{subfigure}
\caption{(\subref{fig:hw_compare}): Comparison between the SOTA accelerator (Eyeriss) and searched design. Results are EDPs normalized to Eyeriss, and lower is better. (\subref{fig:bo_ablation})-(\subref{fig:lcb_ablation}): Ablation studies on ResNet-K4. Higher results are better. (\subref{fig:bo_ablation}) BO with different surrogate models and acquisition functions. (\subref{fig:lcb_ablation}) LCB acquisition function with different $\lambda$ values.}
\vspace{-.25in}
\label{fig:ablations_and_hwcompare}
\end{figure}

\vspace{-.1in}
\paragraph{Surrogate models and acquisition functions}  There exist
popular variants for both the surrogate models and acquisition functions.
In Figure~\ref{fig:bo_ablation}, we compare the surrogate models of Gaussian
process (GP) with random forest (RF) and the acquisition functions of
expected improvement (EI) and lower confidence bound (LCB).  As shown,
in the transformed feature space, GP generally performs better than RF,
and LCB generally outperforms EI.


\paragraph{Exploration vs. Exploitation} The LCB acquisition function explicitly balances exploration and exploitation with a hyperparameter ${\lambda}$. To further study the impact of exploration vs. exploitation in the use of LCB, we test LCB with different ${\lambda}$ values in Figure~\ref{fig:lcb_ablation}. We find that LCBs with ${\lambda}$ values that are greater than 0.5 provide stable performance in the optimization curves, while LCB with ${\lambda}=0.1$ suffers from insufficient exploration.

\vspace{-.05in}
\subsection{Architectural Insights}
\vspace{-.1in}
To show that our automated design can produce new architectural insights,
we provide a qualitative comparison of Eyeriss with our solution for DQN.
Our design predominantly differs in the shape of the PE array, as well
as in the number of memory buffers used.  Eyeriss allocates the majority
of its local buffer storage for filter weights, which are poorly utilized.
Our design increases buffer utilization by storing multiple inputs and
output elements.

\ignore{
One of the benefits of automated design is the opportunity to gain new
architectural insights. To demonstrate this, we provide a qualitative
comparison of Eyeriss with our solution for DQN. Our design predominantly
differs in the shape of the PE array, as well as how memory buffers are
allocated. Eyeriss allocates the majority of its local buffer storage for
filter weights, but it is poorly utilized. Our design increases buffer
utilization by storing multiple inputs and output elements as well.
}

We can also plug our hardware configuration into the heuristic-based
optimizer from prior work \citep{parashar2019timeloop} and attempt to
find a software mapping.  We do this using the 12$\times$14 PE array from
DQN.  Timeloop's software optimizers are unable to find the same mapping
that we do, with the best result being 52\% worse than the baseline.
This demonstrates the utility of a learned co-design approach that enables
the software optimizer to be robust across different hardware architectures.


\vspace{-.1in}
\section{Related Work}
\vspace{-.1in}


\subsection{Hardware to Optimize DNNs}
\vspace{-.1in}

Accelerators are specialized processors that provide significant performance
and efficiency improvements by targeting specific workloads; they also
typically require significant manual design.  For deep learning, these
primitives are often basic linear algebra subprograms (BLAS).

Prior work has designed specialized hardware to execute BLAS kernels.
Google's TPU~\citep{jouppi2017datacenter} uses large hardware structures
called systolic arrays~\citep{kung1979systolic}, and NVIDIA's GPUs have
tensor cores~\citep{v100}.  DianNao~\citep{chen2014diannao} computes fully
connected layers of a neural network using multiply-accumulate trees, and
its successor, DaDianNao~\citep{chen2014dadiannao}, improves data locality
for large neural networks by tiling processing elements (PEs) around on-chip
eDRAM banks.  A more specialized version, ShiDianNao~\citep{du2015shidiannao},
focuses on dense convolutional neural networks (CNNs), which exhibit regular
computational behavior with high data reuse.  Eyeriss~\citep{chen2016eyeriss}
and Eyeriss v2~\citep{chen2019eyeriss} also focus on CNNs, introducing
a specific dataflow that exploits a reuse pattern exhibited by 2D
convolutions.  To improve performance scaling, Eyeriss v2 uses a more
sophisticated interconnect than its predecessor, and it also introduces
a variant that targets sparse CNNs.  Prior work~\citep{parashar2017scnn,
zhang2016cambricon} has dealt with sparsity by suppressing zero-valued
activations and storing and operating on compressed data.  Many other
domain specific architectures have been proposed to take advantage
of local communication patterns~\citep{farabet2011neuflow}, 3D-stacked
bandwidth memory~\citep{kim2016neurocube, gao2017tetris}, or multi-chip
modules~\citep{shao2019simba}. Recent work~\citep{yang2020interstellar} presents heuristics for automatically
synthesizing hardware using a domain-specific language.

\vspace{-.1in}
\subsection{Software to Optimize DNNs}
\vspace{-.1in}

Software optimizations for reducing the compute and storage requirements
of neural networks include loop blocking (tiling), loop reordering, and
loop unrolling~\citep{whaley1998automatically, mullapudi2016automatically,
bondhugula2008practical}.  Compilers such as TVM~\citep{chen2018learning}
have used learned cost models to optimize execution efficiency.  Similarly,
Timeloop uses a grid or random search to optimize software mappings on a
user-specified hardware architecture~\citep{parashar2019timeloop}.  However,
all previous software optimizers treat hardware as a black box and ignore
interactions between hardware and software.

\ignore{
a black box without explicitly considering the interaction between hardware
and software, such as buffer sizing/utilization and parallelism usage.
}

Ours is the first work that systematically explores the space of both
hardware and software optimizations; this larger search space requires a
more principled search method, which motivates our constrained Bayesian
optimization framework.

\vspace{-.15in}
\section{Conclusion}
\label{sec:conclusion}
\vspace{-.1in}

In this paper, we have cast hardware/software co-design as a Bayesian
optimization problem.  We have shown that standard mechanisms have
difficulty navigating the complex, highly constrained design space,
so we have presented a novel constrained formulation that
allows the optimizer to efficiently identify desirable points in this
design space.  The use of machine learning to automate hardware/software
co-design opens many opportunities for future work.  For example, transfer
learning could dramatically reduce design time across designs and models.
The techniques described here are not limited to DNN architectures,
which is significant because as we enter the golden age of computer
architecture~\citep{hennessy2019new}, it is essential that we develop
automatic mechanisms for architectural exploration that quickly produce
custom hardware accelerators.


\bibliography{iclr2021_conference}

\begin{thebibliography}{41}
\providecommand{\natexlab}[1]{#1}
\providecommand{\url}[1]{\texttt{#1}}
\expandafter\ifx\csname urlstyle\endcsname\relax
  \providecommand{\doi}[1]{doi: #1}\else
  \providecommand{\doi}{doi: \begingroup \urlstyle{rm}\Url}\fi

\bibitem[Bondhugula et~al.(2008)Bondhugula, Hartono, Ramanujam, and
  Sadayappan]{bondhugula2008practical}
Uday Bondhugula, Albert Hartono, Jagannathan Ramanujam, and Ponnuswamy
  Sadayappan.
\newblock A practical automatic polyhedral parallelizer and locality optimizer.
\newblock In \emph{Proceedings of the 29th ACM SIGPLAN Conference on
  Programming Language Design and Implementation}, pp.\  101--113, 2008.

\bibitem[Brochu et~al.(2010)Brochu, Cora, and De~Freitas]{brochu2010tutorial}
Eric Brochu, Vlad~M Cora, and Nando De~Freitas.
\newblock A tutorial on bayesian optimization of expensive cost functions, with
  application to active user modeling and hierarchical reinforcement learning.
\newblock \emph{arXiv preprint arXiv:1012.2599}, 2010.

\bibitem[Chen et~al.(2018)Chen, Zheng, Yan, Jiang, Moreau, Ceze, Guestrin, and
  Krishnamurthy]{chen2018learning}
Tianqi Chen, Lianmin Zheng, Eddie Yan, Ziheng Jiang, Thierry Moreau, Luis Ceze,
  Carlos Guestrin, and Arvind Krishnamurthy.
\newblock Learning to optimize tensor programs.
\newblock In \emph{Advances in Neural Information Processing Systems}, pp.\
  3389--3400, 2018.

\bibitem[Chen et~al.(2014{\natexlab{a}})Chen, Du, Sun, Wang, Wu, Chen, and
  Temam]{chen2014diannao}
Tianshi Chen, Zidong Du, Ninghui Sun, Jia Wang, Chengyong Wu, Yunji Chen, and
  Olivier Temam.
\newblock Diannao: a small-footprint high-throughput accelerator for ubiquitous
  machine-learning.
\newblock In \emph{Proceedings of the 19th international conference on
  Architectural Support for Programming Languages and Operating Systems
  (ASPLOS)}, pp.\  269--284, 2014{\natexlab{a}}.

\bibitem[Chen et~al.(2016)Chen, Emer, and Sze]{chen2016eyeriss}
Yu-Hsin Chen, Joel Emer, and Vivienne Sze.
\newblock Eyeriss: A spatial architecture for energy-efficient dataflow for
  convolutional neural networks.
\newblock \emph{ACM SIGARCH Computer Architecture News}, 44\penalty0
  (3):\penalty0 367--379, 2016.

\bibitem[Chen et~al.(2019)Chen, Yang, Emer, and Sze]{chen2019eyeriss}
Yu-Hsin Chen, Tien-Ju Yang, Joel Emer, and Vivienne Sze.
\newblock Eyeriss v2: A flexible accelerator for emerging deep neural networks
  on mobile devices.
\newblock \emph{IEEE Journal on Emerging and Selected Topics in Circuits and
  Systems}, 9\penalty0 (2):\penalty0 292--308, 2019.

\bibitem[Chen et~al.(2014{\natexlab{b}})Chen, Luo, Liu, Zhang, He, Wang, Li,
  Chen, Xu, Sun, et~al.]{chen2014dadiannao}
Yunji Chen, Tao Luo, Shaoli Liu, Shijin Zhang, Liqiang He, Jia Wang, Ling Li,
  Tianshi Chen, Zhiwei Xu, Ninghui Sun, et~al.
\newblock Dadiannao: A machine-learning supercomputer.
\newblock In \emph{2014 47th Annual IEEE/ACM International Symposium on
  Microarchitecture (MICRO)}, pp.\  609--622. IEEE, 2014{\natexlab{b}}.

\bibitem[Du et~al.(2015)Du, Fasthuber, Chen, Ienne, Li, Luo, Feng, Chen, and
  Temam]{du2015shidiannao}
Zidong Du, Robert Fasthuber, Tianshi Chen, Paolo Ienne, Ling Li, Tao Luo,
  Xiaobing Feng, Yunji Chen, and Olivier Temam.
\newblock Shidiannao: Shifting vision processing closer to the sensor.
\newblock In \emph{Proceedings of the 42nd Annual International Symposium on
  Computer Architecture (ISCA)}, pp.\  92--104, 2015.

\bibitem[Farabet et~al.(2011)Farabet, Martini, Corda, Akselrod, Culurciello,
  and LeCun]{farabet2011neuflow}
Cl{\'e}ment Farabet, Berin Martini, Benoit Corda, Polina Akselrod, Eugenio
  Culurciello, and Yann LeCun.
\newblock Neuflow: A runtime reconfigurable dataflow processor for vision.
\newblock In \emph{Cvpr 2011 Workshops}, pp.\  109--116. IEEE, 2011.

\bibitem[Frazier(2009)]{frazier2009knowledge}
Peter~I Frazier.
\newblock \emph{Knowledge-gradient methods for statistical learning}.
\newblock PhD thesis, Citeseer, 2009.

\bibitem[Gao et~al.(2017)Gao, Pu, Yang, Horowitz, and Kozyrakis]{gao2017tetris}
Mingyu Gao, Jing Pu, Xuan Yang, Mark Horowitz, and Christos Kozyrakis.
\newblock Tetris: Scalable and efficient neural network acceleration with 3d
  memory.
\newblock In \emph{Proceedings of the Twenty-Second International Conference on
  Architectural Support for Programming Languages and Operating Systems}, pp.\
  751--764, 2017.

\bibitem[Gelbart et~al.(2014)Gelbart, Snoek, and Adams]{gelbart2014bayesian}
Michael~A Gelbart, Jasper Snoek, and Ryan~P Adams.
\newblock Bayesian optimization with unknown constraints.
\newblock \emph{Uncertainty in Artificial Intelligence}, 2014.

\bibitem[He et~al.(2016)He, Zhang, Ren, and Sun]{he2016deep}
Kaiming He, Xiangyu Zhang, Shaoqing Ren, and Jian Sun.
\newblock Deep residual learning for image recognition.
\newblock In \emph{Proceedings of the IEEE conference on computer vision and
  pattern recognition}, pp.\  770--778, 2016.

\bibitem[Hennessy \& Patterson(2019)Hennessy and Patterson]{hennessy2019new}
John~L Hennessy and David~A Patterson.
\newblock A new golden age for computer architecture.
\newblock \emph{Communications of the ACM}, 62\penalty0 (2):\penalty0 48--60,
  2019.

\bibitem[Hennig \& Schuler(2012)Hennig and Schuler]{hennig2012entropy}
Philipp Hennig and Christian~J Schuler.
\newblock Entropy search for information-efficient global optimization.
\newblock \emph{Journal of Machine Learning Research}, 13\penalty0
  (Jun):\penalty0 1809--1837, 2012.

\bibitem[Hernandez \& Brown(2020)Hernandez and Brown]{hern2020measuring}
Danny Hernandez and Tom~B. Brown.
\newblock Measuring the algorithmic efficiency of neural networks, 2020.

\bibitem[Hern{\'a}ndez-Lobato et~al.(2014)Hern{\'a}ndez-Lobato, Hoffman, and
  Ghahramani]{hernandez2014predictive}
Jos{\'e}~Miguel Hern{\'a}ndez-Lobato, Matthew~W Hoffman, and Zoubin Ghahramani.
\newblock Predictive entropy search for efficient global optimization of
  black-box functions.
\newblock In \emph{Advances in neural information processing systems}, pp.\
  918--926, 2014.

\bibitem[Hutter et~al.(2011)Hutter, Hoos, and
  Leyton-Brown]{hutter2011sequential}
Frank Hutter, Holger~H Hoos, and Kevin Leyton-Brown.
\newblock Sequential model-based optimization for general algorithm
  configuration.
\newblock In \emph{International conference on learning and intelligent
  optimization}, pp.\  507--523. Springer, 2011.

\bibitem[Jones et~al.(1998)Jones, Schonlau, and Welch]{jones1998efficient}
Donald~R Jones, Matthias Schonlau, and William~J Welch.
\newblock Efficient global optimization of expensive black-box functions.
\newblock \emph{Journal of Global optimization}, 13\penalty0 (4):\penalty0
  455--492, 1998.

\bibitem[Jouppi et~al.(2017)Jouppi, Young, Patil, Patterson, Agrawal, Bajwa,
  Bates, Bhatia, Boden, Borchers, et~al.]{jouppi2017datacenter}
Norman~P Jouppi, Cliff Young, Nishant Patil, David Patterson, Gaurav Agrawal,
  Raminder Bajwa, Sarah Bates, Suresh Bhatia, Nan Boden, Al~Borchers, et~al.
\newblock In-datacenter performance analysis of a tensor processing unit.
\newblock In \emph{Proceedings of the 44th Annual International Symposium on
  Computer Architecture (ISCA)}, pp.\  1--12, 2017.

\bibitem[Kim et~al.(2016)Kim, Kung, Chai, Yalamanchili, and
  Mukhopadhyay]{kim2016neurocube}
Duckhwan Kim, Jaeha Kung, Sek Chai, Sudhakar Yalamanchili, and Saibal
  Mukhopadhyay.
\newblock Neurocube: A programmable digital neuromorphic architecture with
  high-density 3d memory.
\newblock \emph{ACM SIGARCH Computer Architecture News}, 44\penalty0
  (3):\penalty0 380--392, 2016.

\bibitem[Kingma \& Ba(2014)Kingma and Ba]{kingma2014adam}
Diederik~P Kingma and Jimmy Ba.
\newblock Adam: A method for stochastic optimization.
\newblock \emph{arXiv preprint arXiv:1412.6980}, 2014.

\bibitem[Kung \& Leiserson(1979)Kung and Leiserson]{kung1979systolic}
HT~Kung and Charles~E Leiserson.
\newblock Systolic arrays (for vlsi).
\newblock In \emph{Sparse Matrix Proceedings 1978}, volume~1, pp.\  256--282.
  Society for industrial and applied mathematics, 1979.

\bibitem[Letham et~al.(2019)Letham, Karrer, Ottoni, Bakshy,
  et~al.]{letham2019constrained}
Benjamin Letham, Brian Karrer, Guilherme Ottoni, Eytan Bakshy, et~al.
\newblock Constrained bayesian optimization with noisy experiments.
\newblock \emph{Bayesian Analysis}, 14\penalty0 (2):\penalty0 495--519, 2019.

\bibitem[Mnih et~al.(2013)Mnih, Kavukcuoglu, Silver, Graves, Antonoglou,
  Wierstra, and Riedmiller]{mnih2013playing}
Volodymyr Mnih, Koray Kavukcuoglu, David Silver, Alex Graves, Ioannis
  Antonoglou, Daan Wierstra, and Martin Riedmiller.
\newblock Playing atari with deep reinforcement learning.
\newblock \emph{arXiv preprint arXiv:1312.5602}, 2013.

\bibitem[Mullapudi et~al.(2016)Mullapudi, Adams, Sharlet, Ragan-Kelley, and
  Fatahalian]{mullapudi2016automatically}
Ravi~Teja Mullapudi, Andrew Adams, Dillon Sharlet, Jonathan Ragan-Kelley, and
  Kayvon Fatahalian.
\newblock Automatically scheduling halide image processing pipelines.
\newblock \emph{ACM Transactions on Graphics (TOG)}, 35\penalty0 (4):\penalty0
  1--11, 2016.

\bibitem[Nardi et~al.(2019)Nardi, Koeplinger, and Olukotun]{nardi2019practical}
Luigi Nardi, David Koeplinger, and Kunle Olukotun.
\newblock Practical design space exploration.
\newblock In \emph{2019 IEEE 27th International Symposium on Modeling,
  Analysis, and Simulation of Computer and Telecommunication Systems
  (MASCOTS)}, pp.\  347--358. IEEE, 2019.

\bibitem[{NVIDIA}(2017)]{v100}
{NVIDIA}.
\newblock {NVIDIA Tesla V100 GPU Architecture, The World’s Most Advanced Data
  Center GPU}.
\newblock \emph{NVIDIA Corporation}, 2017.

\bibitem[Parashar et~al.(2017)Parashar, Rhu, Mukkara, Puglielli, Venkatesan,
  Khailany, Emer, Keckler, and Dally]{parashar2017scnn}
Angshuman Parashar, Minsoo Rhu, Anurag Mukkara, Antonio Puglielli, Rangharajan
  Venkatesan, Brucek Khailany, Joel Emer, Stephen~W Keckler, and William~J
  Dally.
\newblock Scnn: An accelerator for compressed-sparse convolutional neural
  networks.
\newblock In \emph{Proceedings of the 44th Annual International Symposium on
  Computer Architecture (ISCA)}, pp.\  27--40, 2017.

\bibitem[Parashar et~al.(2019)Parashar, Raina, Shao, Chen, Ying, Mukkara,
  Venkatesan, Khailany, Keckler, and Emer]{parashar2019timeloop}
Angshuman Parashar, Priyanka Raina, Yakun~Sophia Shao, Yu-Hsin Chen, Victor~A
  Ying, Anurag Mukkara, Rangharajan Venkatesan, Brucek Khailany, Stephen~W
  Keckler, and Joel Emer.
\newblock Timeloop: A systematic approach to dnn accelerator evaluation.
\newblock In \emph{2019 IEEE International Symposium on Performance Analysis of
  Systems and Software (ISPASS)}, pp.\  304--315. IEEE, 2019.

\bibitem[Rasmussen \& Williams(2006)Rasmussen and Williams]{gpml2006book}
Carl~Edward Rasmussen and Christopher~K.I. Williams.
\newblock \emph{Gaussian Processes for Machine Learning}.
\newblock MIT Press, 2006.

\bibitem[Shahriari et~al.(2015)Shahriari, Swersky, Wang, Adams, and
  De~Freitas]{shahriari2015taking}
Bobak Shahriari, Kevin Swersky, Ziyu Wang, Ryan~P Adams, and Nando De~Freitas.
\newblock Taking the human out of the loop: A review of bayesian optimization.
\newblock \emph{Proceedings of the IEEE}, 104\penalty0 (1):\penalty0 148--175,
  2015.

\bibitem[Shao et~al.(2019)Shao, Clemons, Venkatesan, Zimmer, Fojtik, Jiang,
  Keller, Klinefelter, Pinckney, and Raina]{shao2019simba}
Yakun~Sophia Shao, Jason Clemons, Rangharajan Venkatesan, Brian Zimmer, Matthew
  Fojtik, Nan Jiang, Ben Keller, Alicia Klinefelter, Nathaniel Pinckney, and
  Priyanka Raina.
\newblock Simba: Scaling deep-learning inference with multi-chip-module-based
  architecture.
\newblock In \emph{Proceedings of the 52nd Annual IEEE/ACM International
  Symposium on Microarchitecture (MICRO)}, 2019.

\bibitem[Snoek et~al.(2012)Snoek, Larochelle, and Adams]{snoek2012practical}
Jasper Snoek, Hugo Larochelle, and Ryan~P Adams.
\newblock Practical bayesian optimization of machine learning algorithms.
\newblock In \emph{Advances in neural information processing systems}, pp.\
  2951--2959, 2012.

\bibitem[Srinivas et~al.(2009)Srinivas, Krause, Kakade, and
  Seeger]{srinivas2009gaussian}
Niranjan Srinivas, Andreas Krause, Sham~M Kakade, and Matthias Seeger.
\newblock Gaussian process optimization in the bandit setting: No regret and
  experimental design.
\newblock \emph{arXiv preprint arXiv:0912.3995}, 2009.

\bibitem[Tan \& Le(2019)Tan and Le]{tan2019efficientnet}
Mingxing Tan and Quoc~V. Le.
\newblock Efficientnet: Rethinking model scaling for convolutional neural
  networks, 2019.

\bibitem[Thompson(1933)]{thompson1933likelihood}
William~R Thompson.
\newblock On the likelihood that one unknown probability exceeds another in
  view of the evidence of two samples.
\newblock \emph{Biometrika}, 25\penalty0 (3/4):\penalty0 285--294, 1933.

\bibitem[Vaswani et~al.(2017)Vaswani, Shazeer, Parmar, Uszkoreit, Jones, Gomez,
  Kaiser, and Polosukhin]{vaswani2017attention}
Ashish Vaswani, Noam Shazeer, Niki Parmar, Jakob Uszkoreit, Llion Jones,
  Aidan~N Gomez, {\L}ukasz Kaiser, and Illia Polosukhin.
\newblock Attention is all you need.
\newblock In \emph{Advances in neural information processing systems}, pp.\
  5998--6008, 2017.

\bibitem[Whaley \& Dongarra(1998)Whaley and Dongarra]{whaley1998automatically}
R~Clinton Whaley and Jack~J Dongarra.
\newblock Automatically tuned linear algebra software.
\newblock In \emph{SC'98: Proceedings of the 1998 ACM/IEEE conference on
  Supercomputing}, pp.\  38--38. IEEE, 1998.

\bibitem[Yang et~al.(2020)Yang, Gao, Liu, Setter, Pu, Nayak, Bell, Cao, Ha,
  Raina, et~al.]{yang2020interstellar}
Xuan Yang, Mingyu Gao, Qiaoyi Liu, Jeff Setter, Jing Pu, Ankita Nayak, Steven
  Bell, Kaidi Cao, Heonjae Ha, Priyanka Raina, et~al.
\newblock Interstellar: Using halide's scheduling language to analyze dnn
  accelerators.
\newblock In \emph{Proceedings of the Twenty-Fifth International Conference on
  Architectural Support for Programming Languages and Operating Systems
  (ASPLOS)}, pp.\  369--383, 2020.

\bibitem[Zhang et~al.(2016)Zhang, Du, Zhang, Lan, Liu, Li, Guo, Chen, and
  Chen]{zhang2016cambricon}
Shijin Zhang, Zidong Du, Lei Zhang, Huiying Lan, Shaoli Liu, Ling Li, Qi~Guo,
  Tianshi Chen, and Yunji Chen.
\newblock Cambricon-x: An accelerator for sparse neural networks.
\newblock In \emph{2016 49th Annual IEEE/ACM International Symposium on
  Microarchitecture (MICRO)}, pp.\  1--12. IEEE, 2016.

\end{thebibliography}
\bibliographystyle{iclr2021_conference}

\appendix
\newpage
\section{Appendix}
\subsection{Parameters and constraints}

\begin{figure}[ht]
  \centering
  \includegraphics[width=0.8\textwidth]{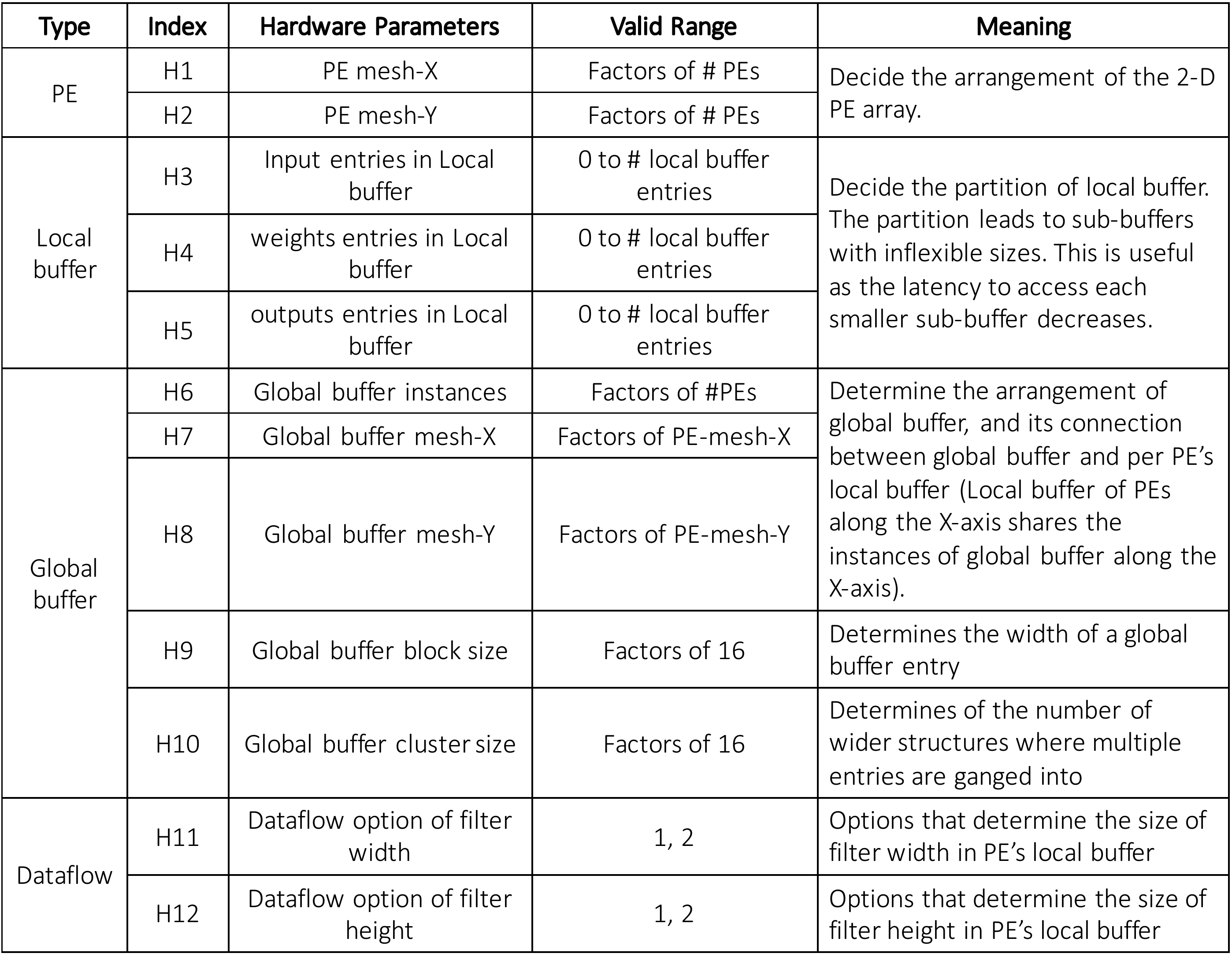}
  \caption{Hardware parameters.}
  \label{fig:hw_params}
\end{figure}

\begin{figure}[ht]
  \centering
  \includegraphics[width=0.8\textwidth]{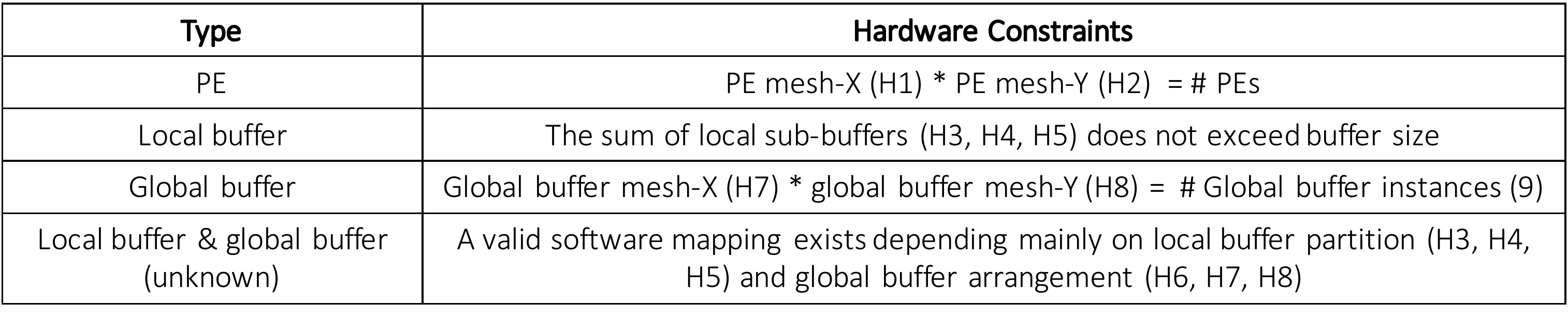}
  \caption{Hardware constraints.}
  \label{fig:hw_cons}
\end{figure}

\begin{figure}[ht]
  \centering
  \includegraphics[width=0.8\textwidth]{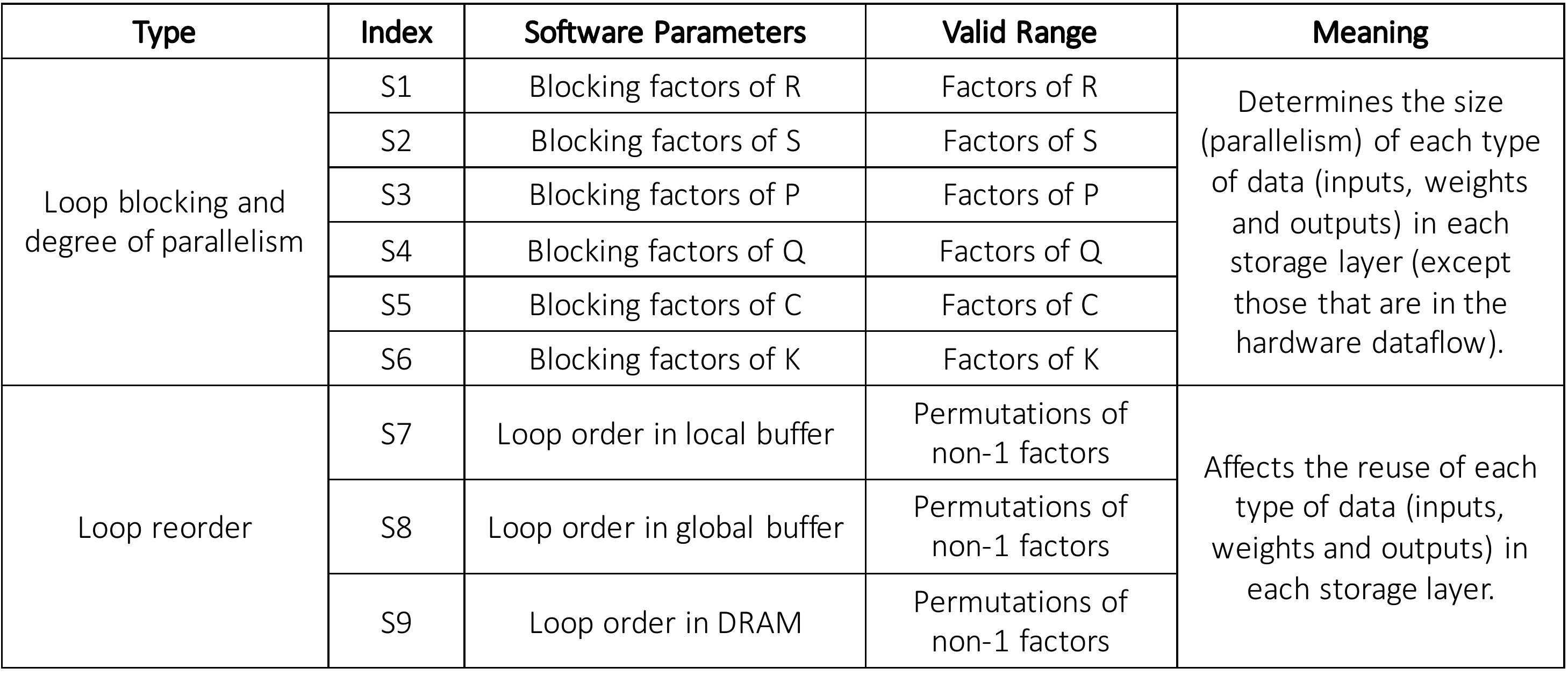}
  \caption{Software parameters.}
  \label{fig:sw_params}
\end{figure}

\begin{figure}[ht]
  \centering
  \includegraphics[width=0.8\textwidth]{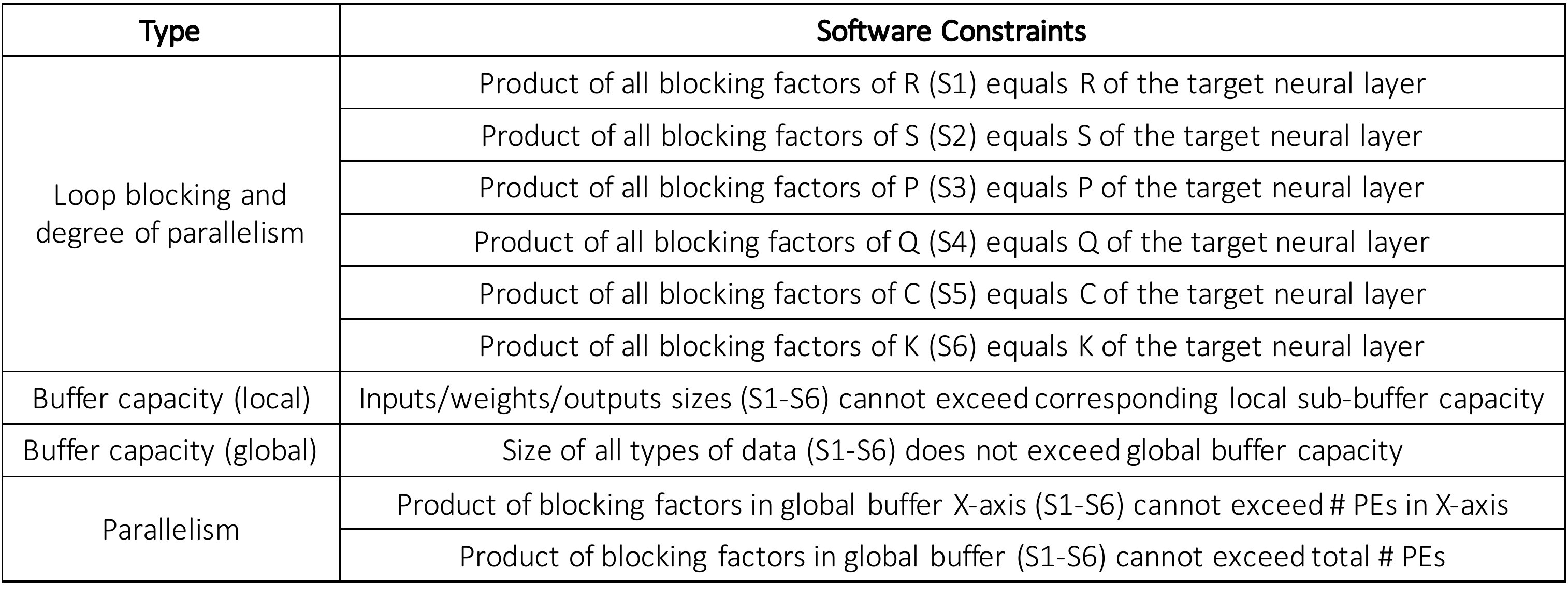}
  \caption{Software constraints.}
  \label{fig:sw_cons}
\end{figure}

\section{Hyperparamters for BO}
\begin{figure}[ht]
  \centering
  \includegraphics[width=0.6\textwidth]{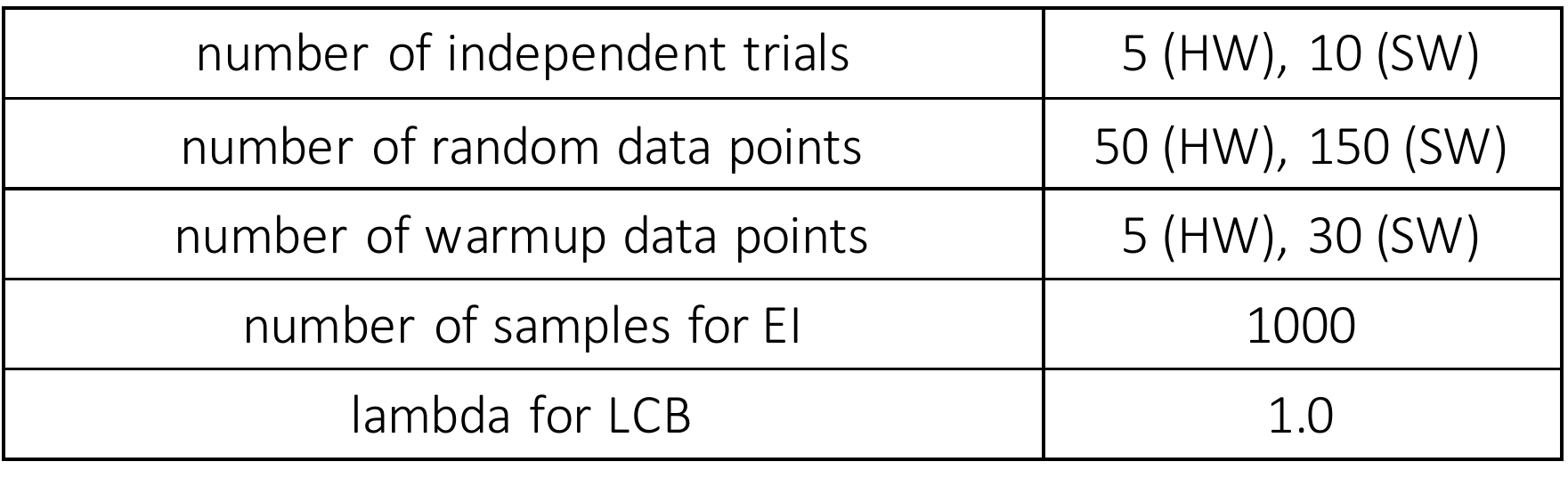}
  \caption{Hyperparamters for BO.}
  \label{fig:hypers}
\end{figure}

In Figure~\ref{fig:hypers} we report the hyperparamters for BO.

\section{Neural Model Specifications.}
\begin{figure}[ht]
  \centering
  \includegraphics[width=0.8\textwidth]{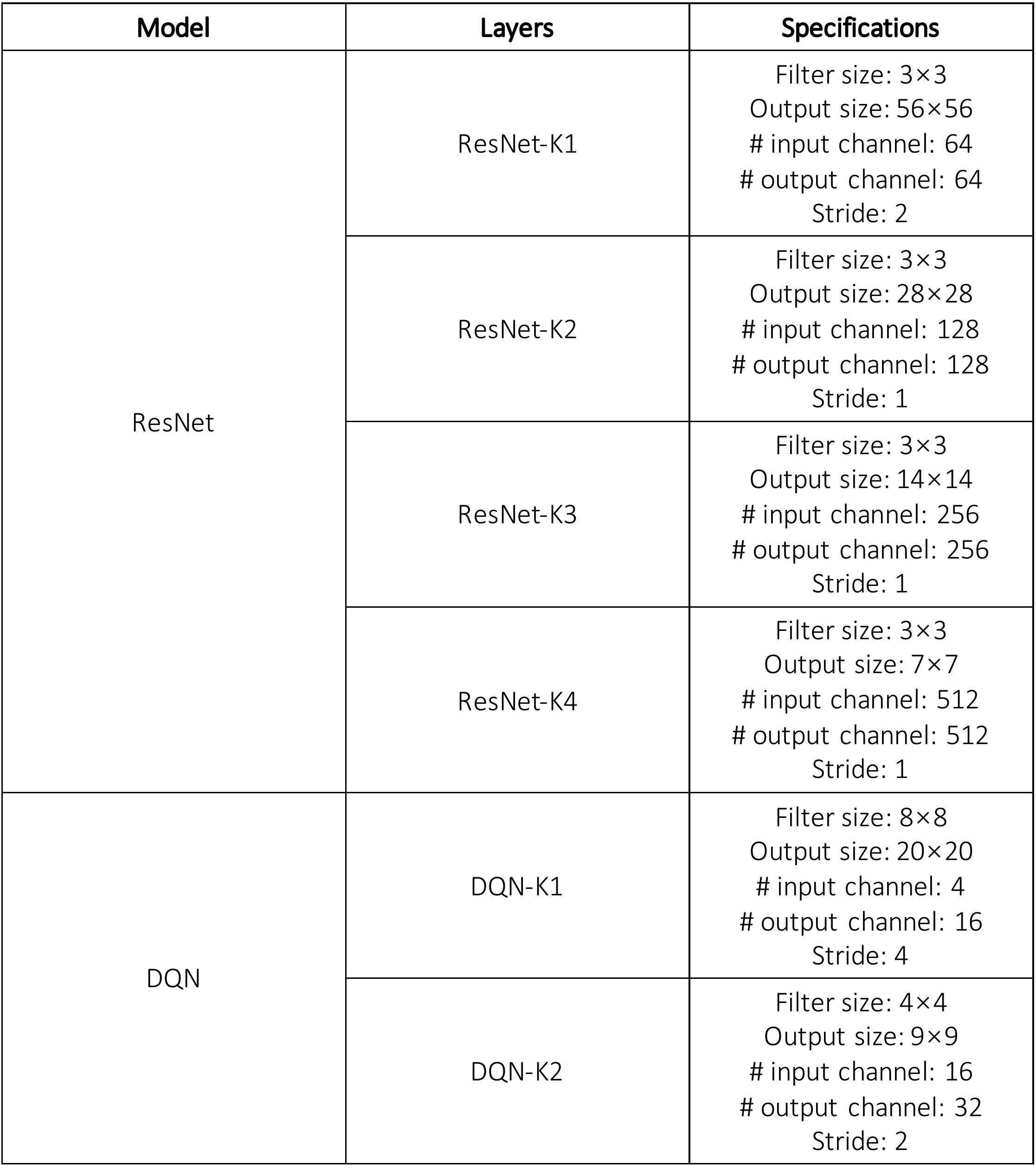}
  \caption{Specifications of ResNet (ResNet-18)~\citep{he2016deep} and DQN~\citep{mnih2013playing}}
  \label{fig:cnn_layers}
\end{figure}

\begin{figure}[ht]
  \centering
  \includegraphics[width=0.8\textwidth]{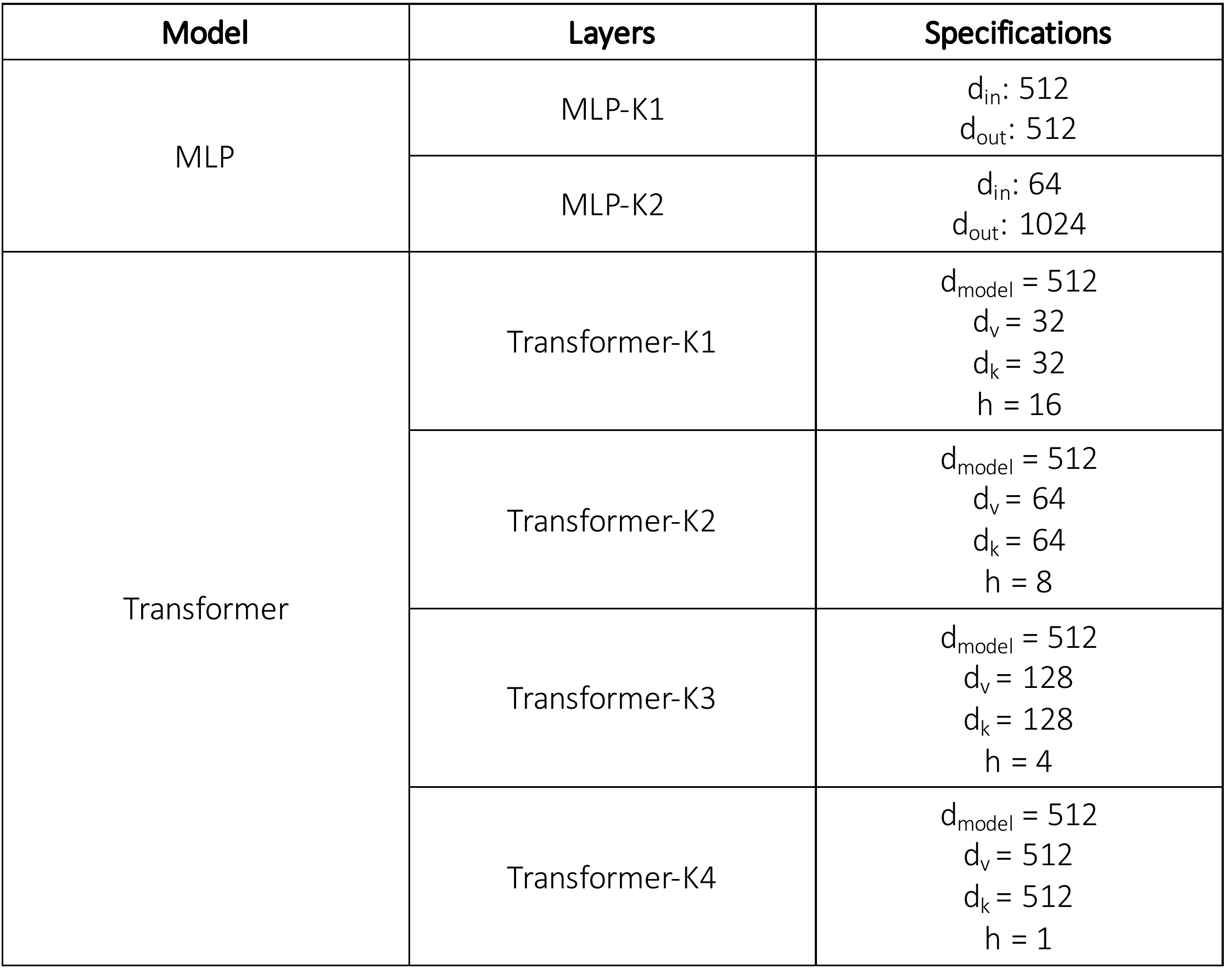}
  \caption{Specifications of MLP and Transformer~\citep{vaswani2017attention}}
  \label{fig:other_layers}
\end{figure}

In Figure~\ref{fig:cnn_layers} and Figure~\ref{fig:other_layers} we report the specifications of neural models benchmarked in this paper.

\begin{figure}[ht]
  \centering
  \includegraphics[width=0.8\textwidth]{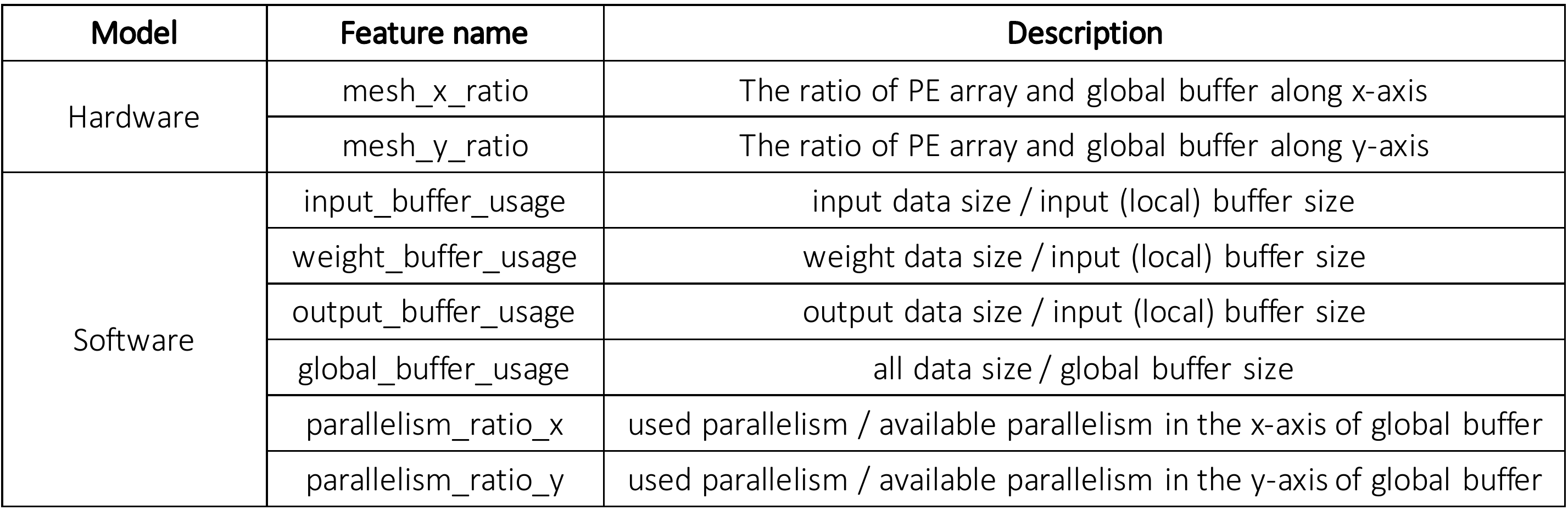}
  \caption{Extra features used by the hardware and software BO optimizers.}
  \label{fig:features}
\end{figure}

\section{Paramterization of 2D Convolution}
Listing~\ref{fig:convloop} gives the seven-level nested loop that comprises a 2D convolution.

\begin{figure}[ht]
 \centering
 \begin{lstlisting}
for n in [0:N)
  for k in [0:K)
    for r in [0:R)
      for s in [0:S)
        for p in [0:P)
          for q in [0:Q)
            for c in [0:C)
              outputs[n][k][q][p] += weights[k][c][s][r] *
                                     inputs[n][c][q+s][p+r]
 \end{lstlisting}
 \caption{Computing a 2D convolution with a seven-level nested loop.}
 \label{fig:convloop}
\end{figure}

\begin{figure}[ht]
  \centering
  \includegraphics[width=\textwidth]{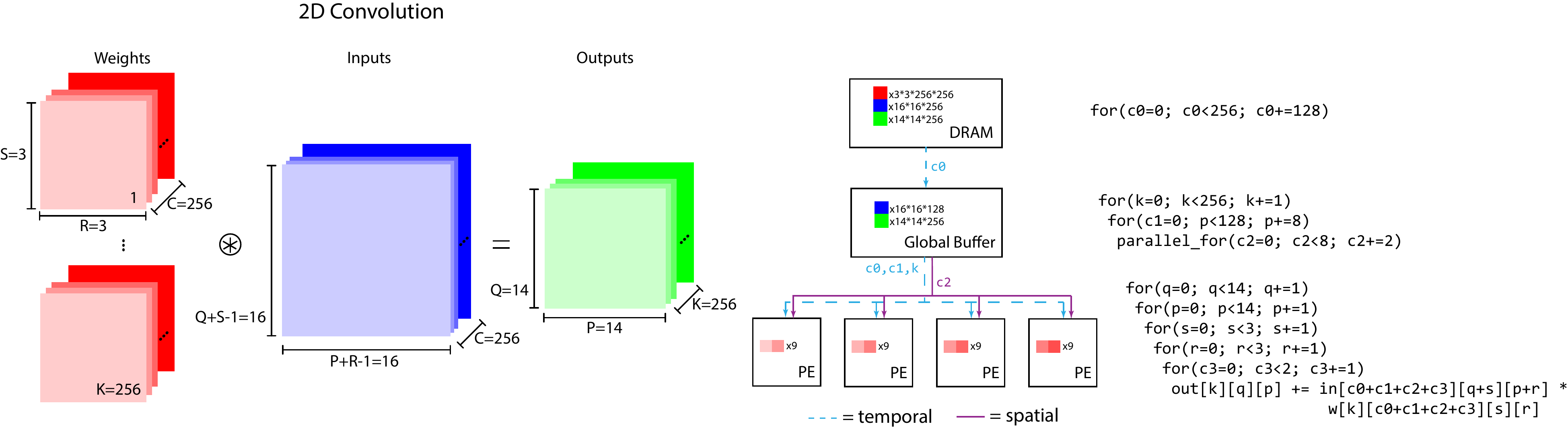}
  \caption{An architecture computing the CONV4 layer of ResNet.}
  \label{fig:conv2d}
\end{figure}

Figure~\ref{fig:conv2d} shows a design point for the CONV4 layer of ResNet. The architecture components are again the same as in the 1D example, but since the memory footprint is significantly larger, the PE can no longer capture all data reuse, so the Global Buffer must store large portions of the inputs and outputs. 

\section{Additional results}
\subsection{Software optimization}
\begin{figure}[ht]
\centering
\begin{subfigure}{.3\textwidth}
  \centering
  \includegraphics[width=1\textwidth]{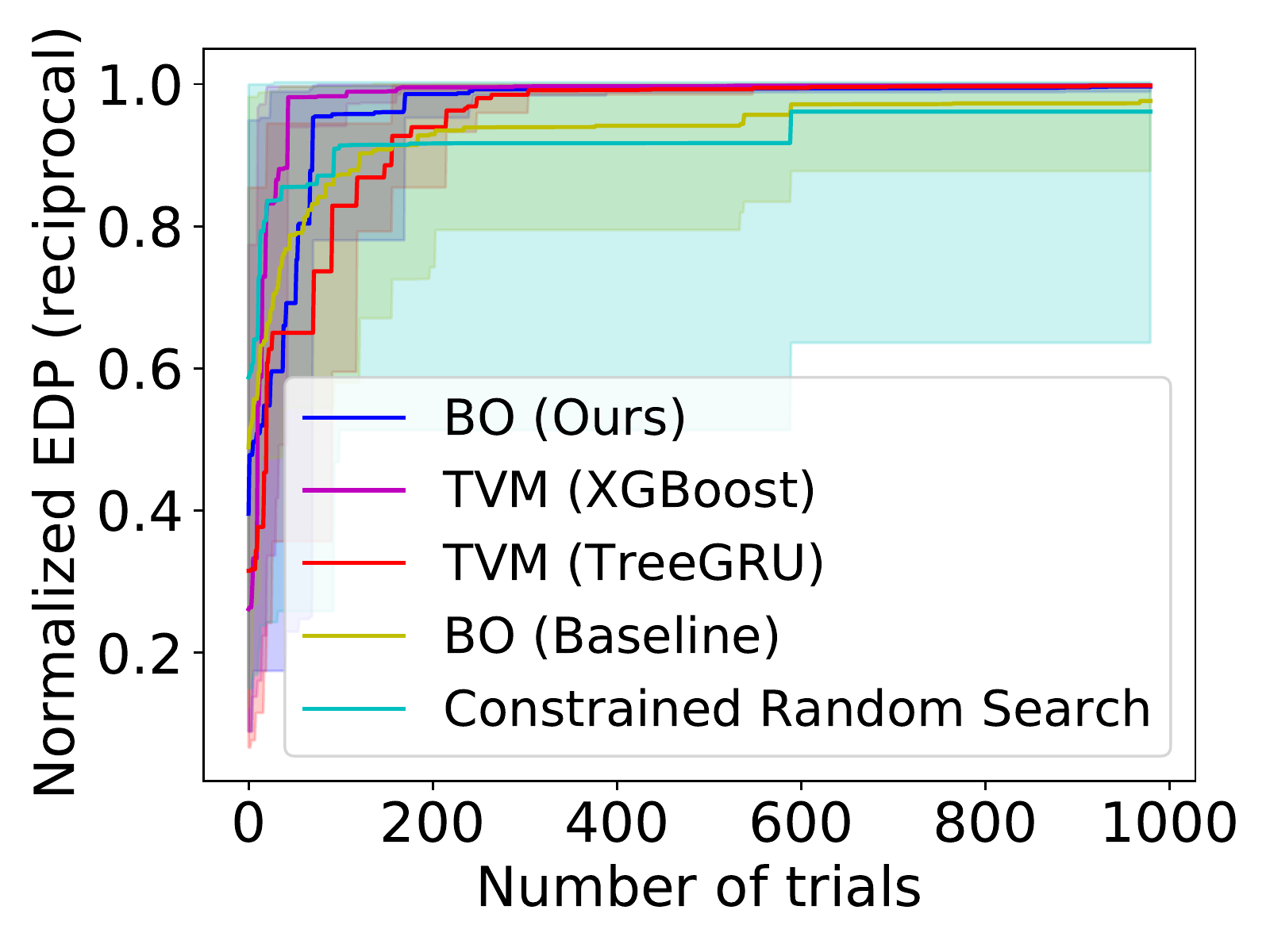}
  \caption{ResNet-K1}
\end{subfigure}
\begin{subfigure}{.3\textwidth}
  \centering
  \includegraphics[width=1\textwidth]{figures/appendix/resnet-2.pdf}
  \caption{ResNet-K2}
\end{subfigure}
\begin{subfigure}{.3\textwidth}
  \centering
  \includegraphics[width=1\textwidth]{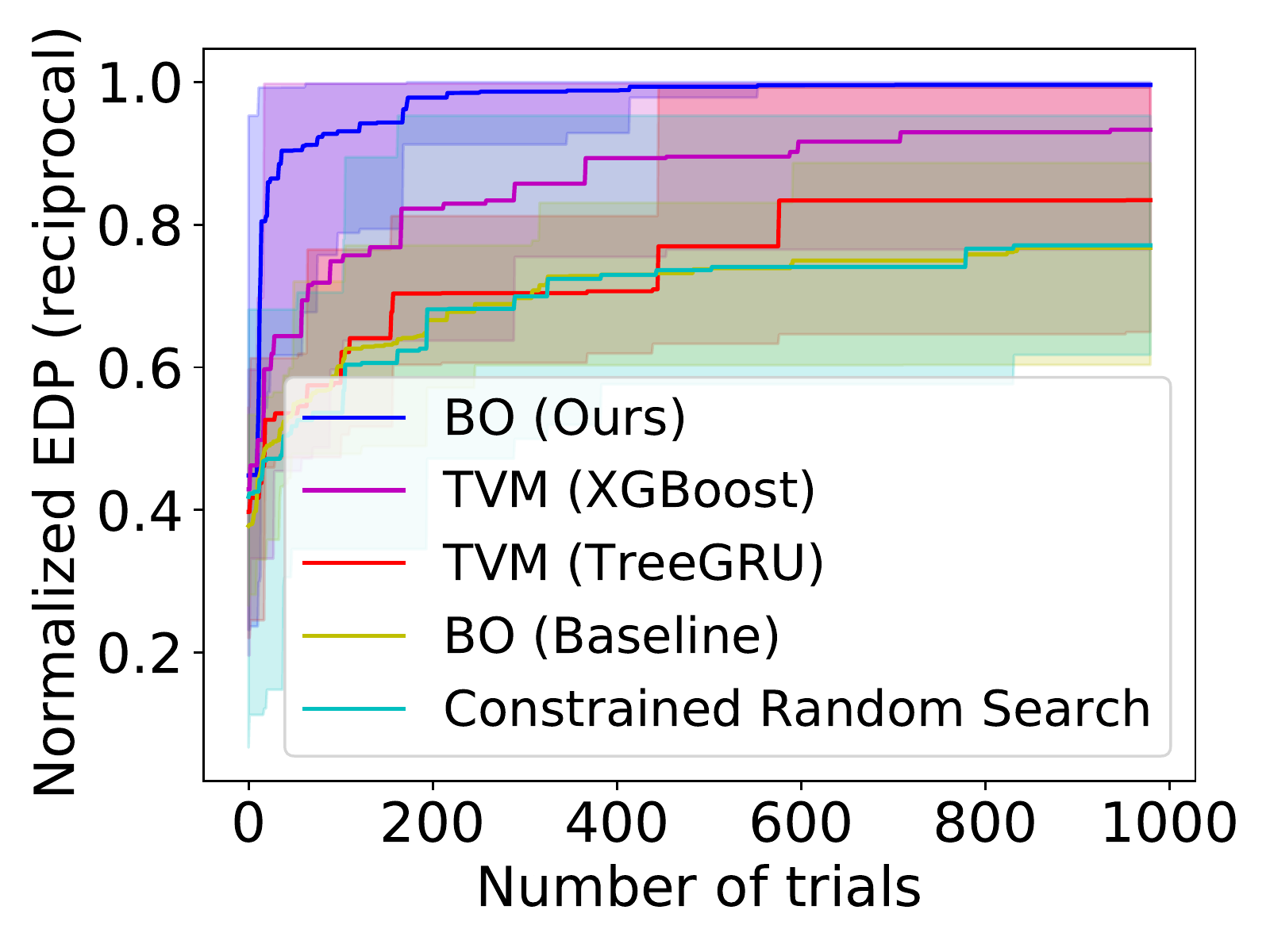}
  \caption{ResNet-K3}
\end{subfigure}
\begin{subfigure}{.3\textwidth}
  \centering
  \includegraphics[width=1\textwidth]{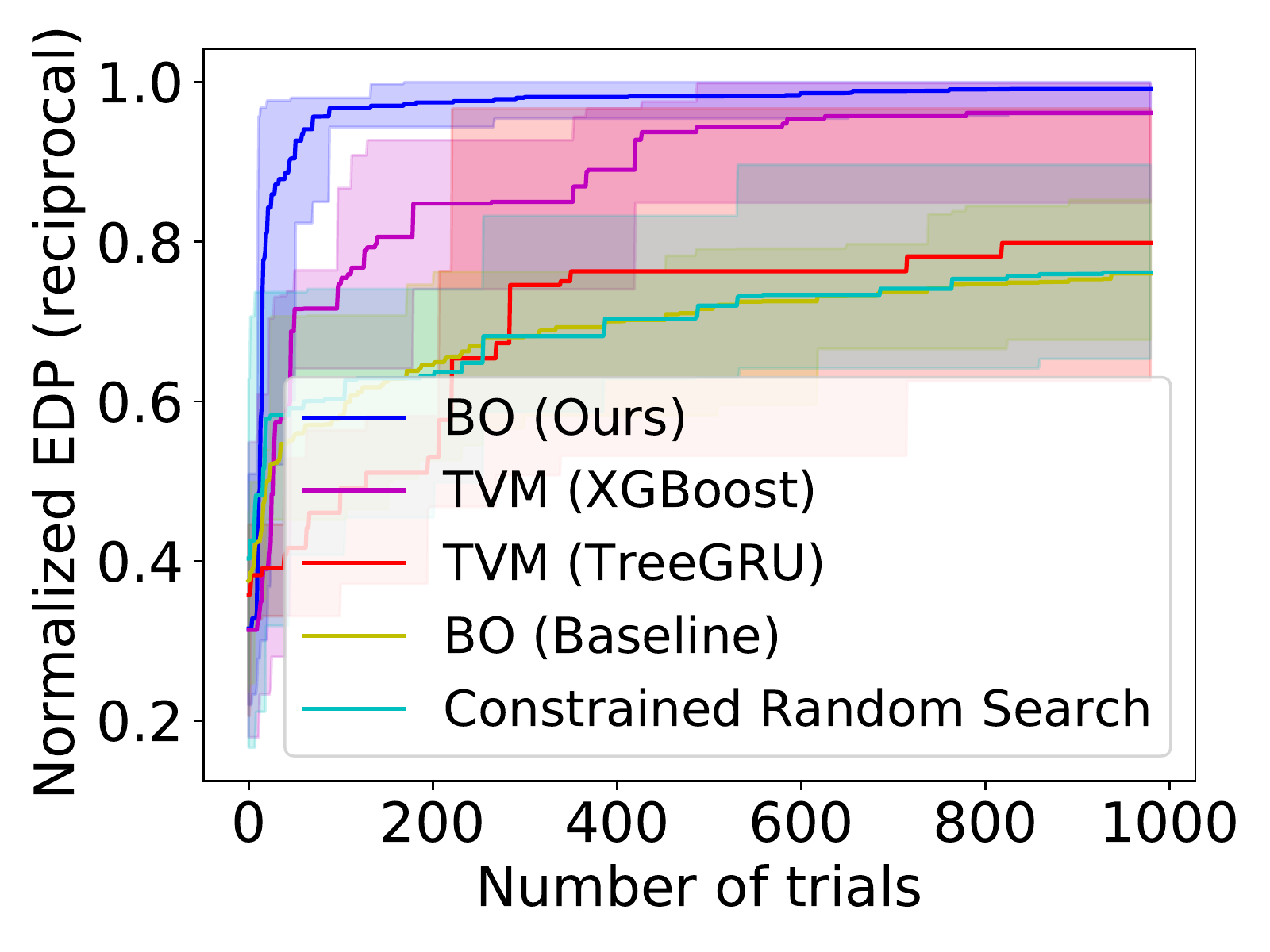}
  \caption{ResNet-K4}
\end{subfigure}
\begin{subfigure}{.3\textwidth}
  \centering
  \includegraphics[width=1\textwidth]{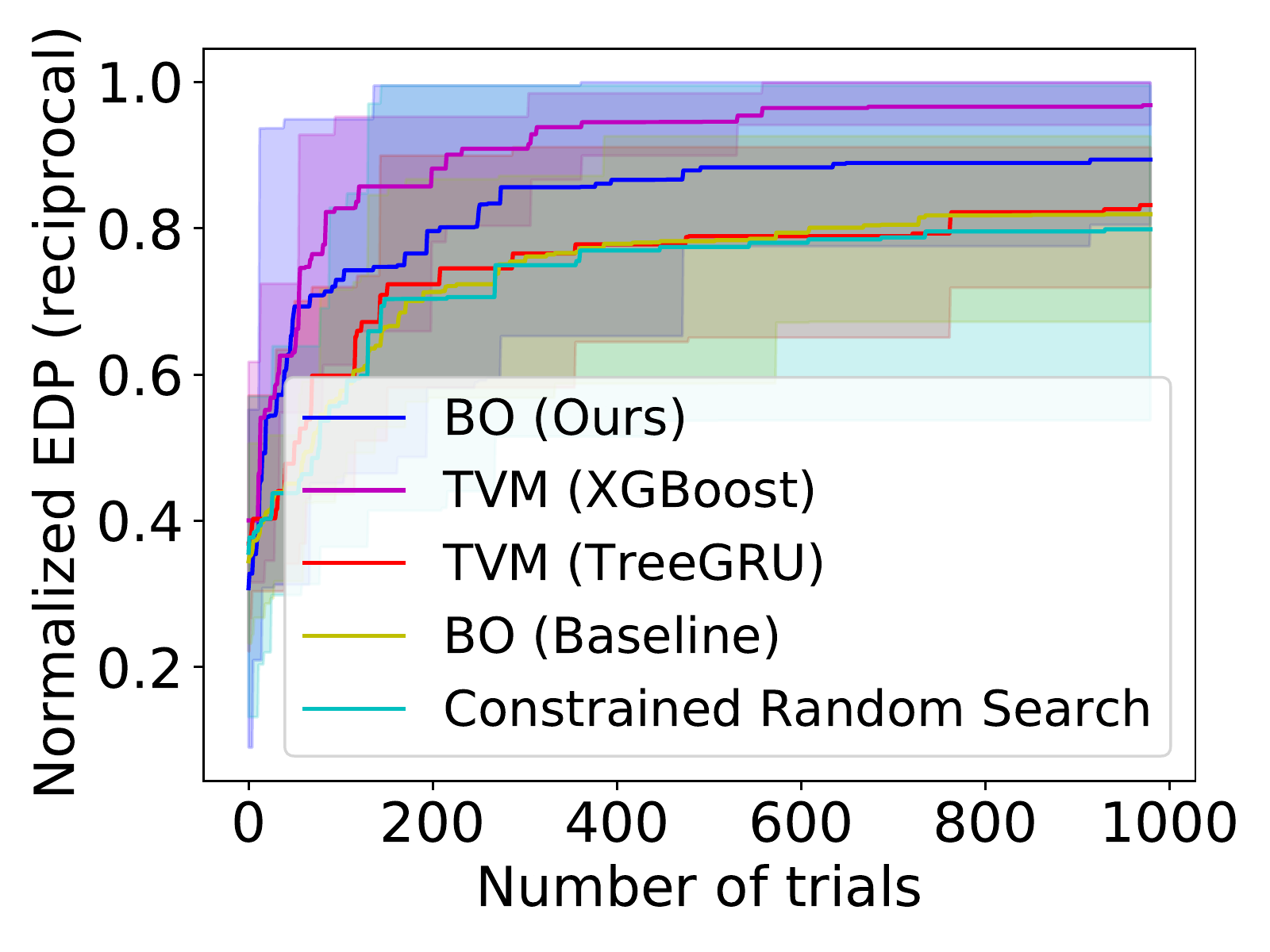}
  \caption{DQN-K1}
\end{subfigure}
\begin{subfigure}{.3\textwidth}
  \centering
  \includegraphics[width=1\textwidth]{figures/appendix/dqn-2.pdf}
  \caption{DQN-K2}
\end{subfigure}
\begin{subfigure}{.3\textwidth}
  \centering
  \includegraphics[width=1\textwidth]{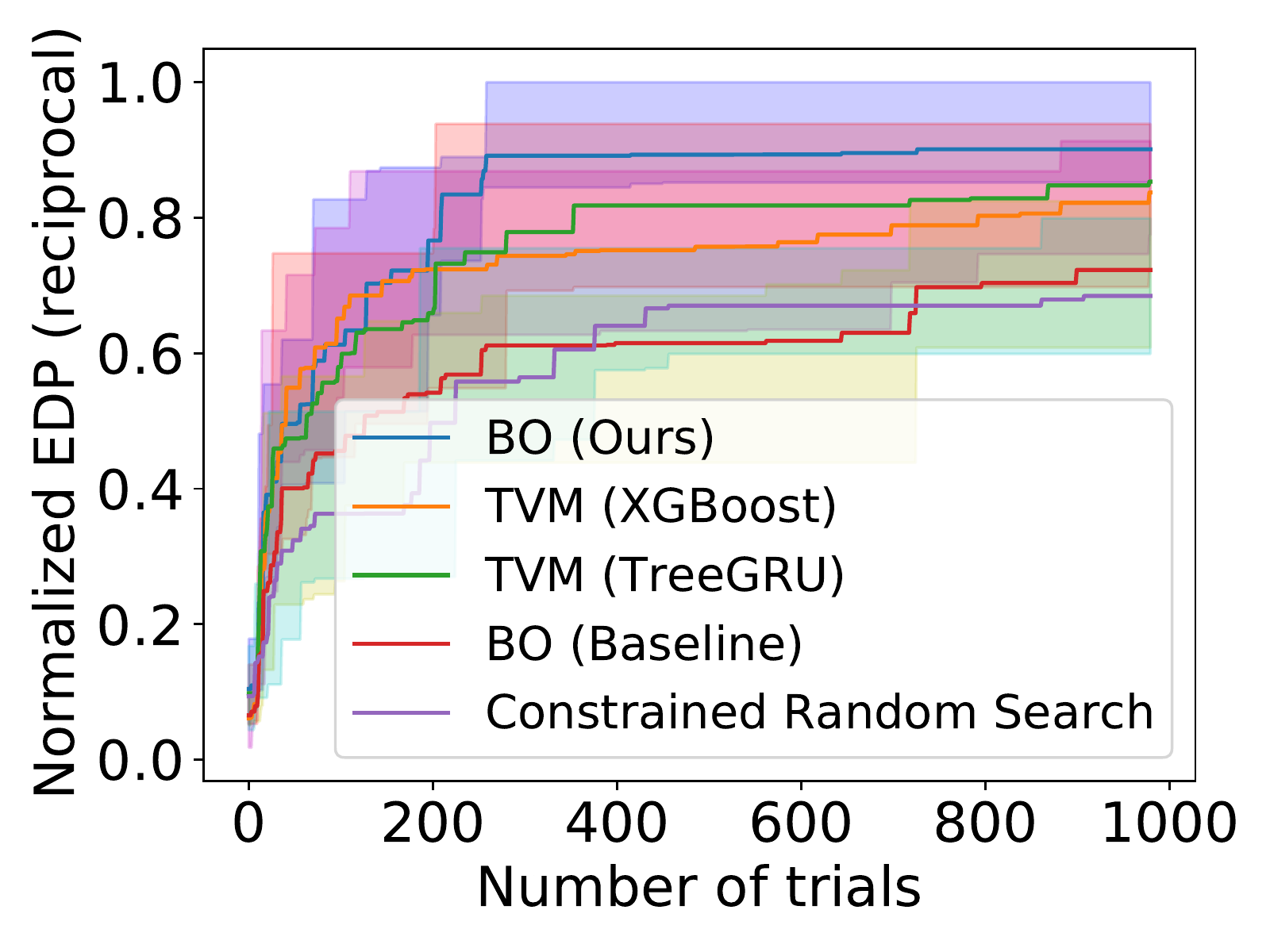}
  \caption{MLP-K1}
\end{subfigure}
\begin{subfigure}{.3\textwidth}
  \centering
  \includegraphics[width=1\textwidth]{figures/appendix/mlp-2.pdf}
  \caption{MLP-K2}
\end{subfigure}
\begin{subfigure}{.3\textwidth}
  \centering
  \includegraphics[width=1\textwidth]{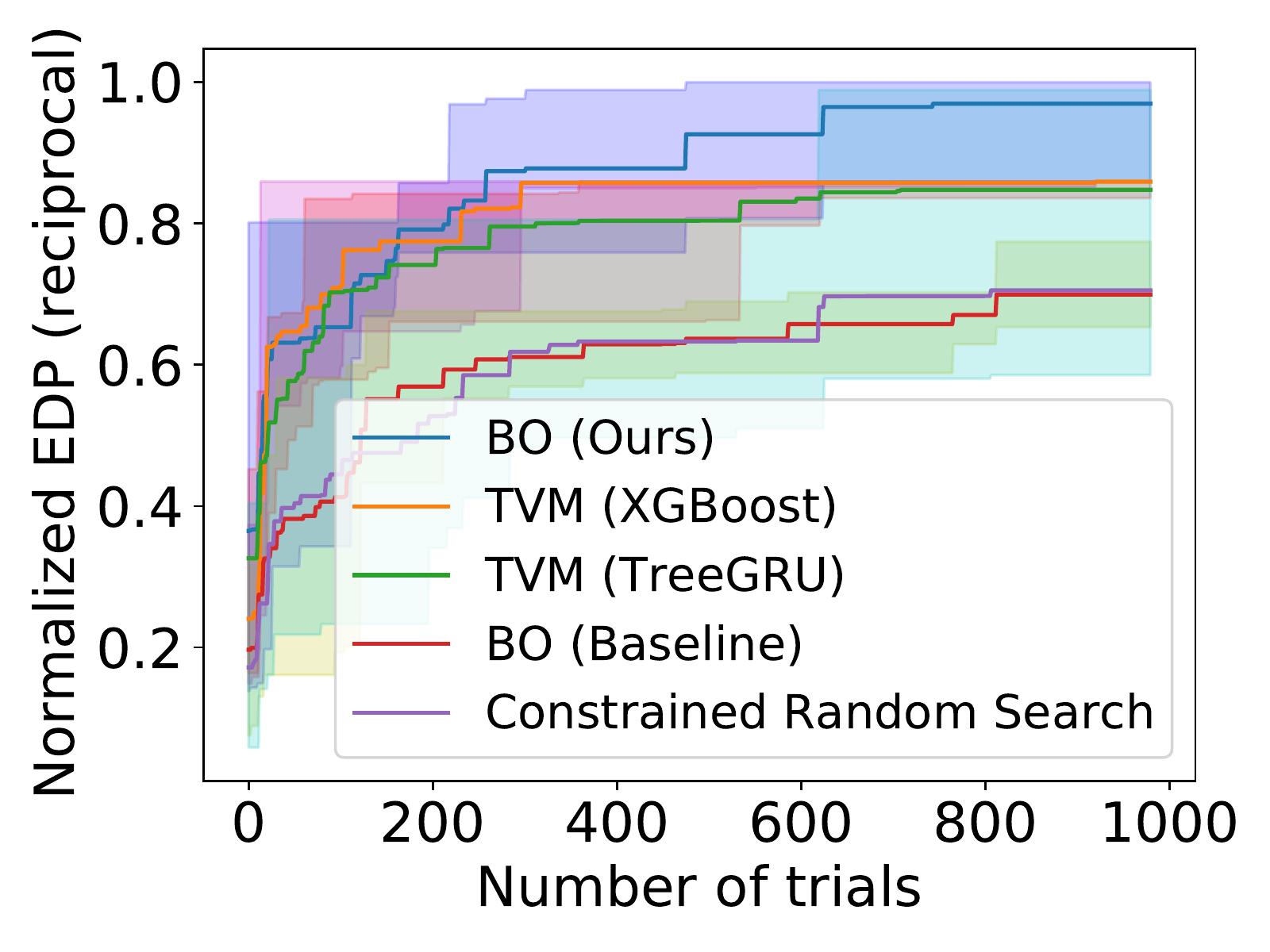}
  \caption{Transformer-K1}
\end{subfigure}
\begin{subfigure}{.3\textwidth}
  \centering
  \includegraphics[width=1\textwidth]{figures/appendix/transformer-2.pdf}
  \caption{Transformer-K2}
\end{subfigure}
\begin{subfigure}{.3\textwidth}
  \centering
  \includegraphics[width=1\textwidth]{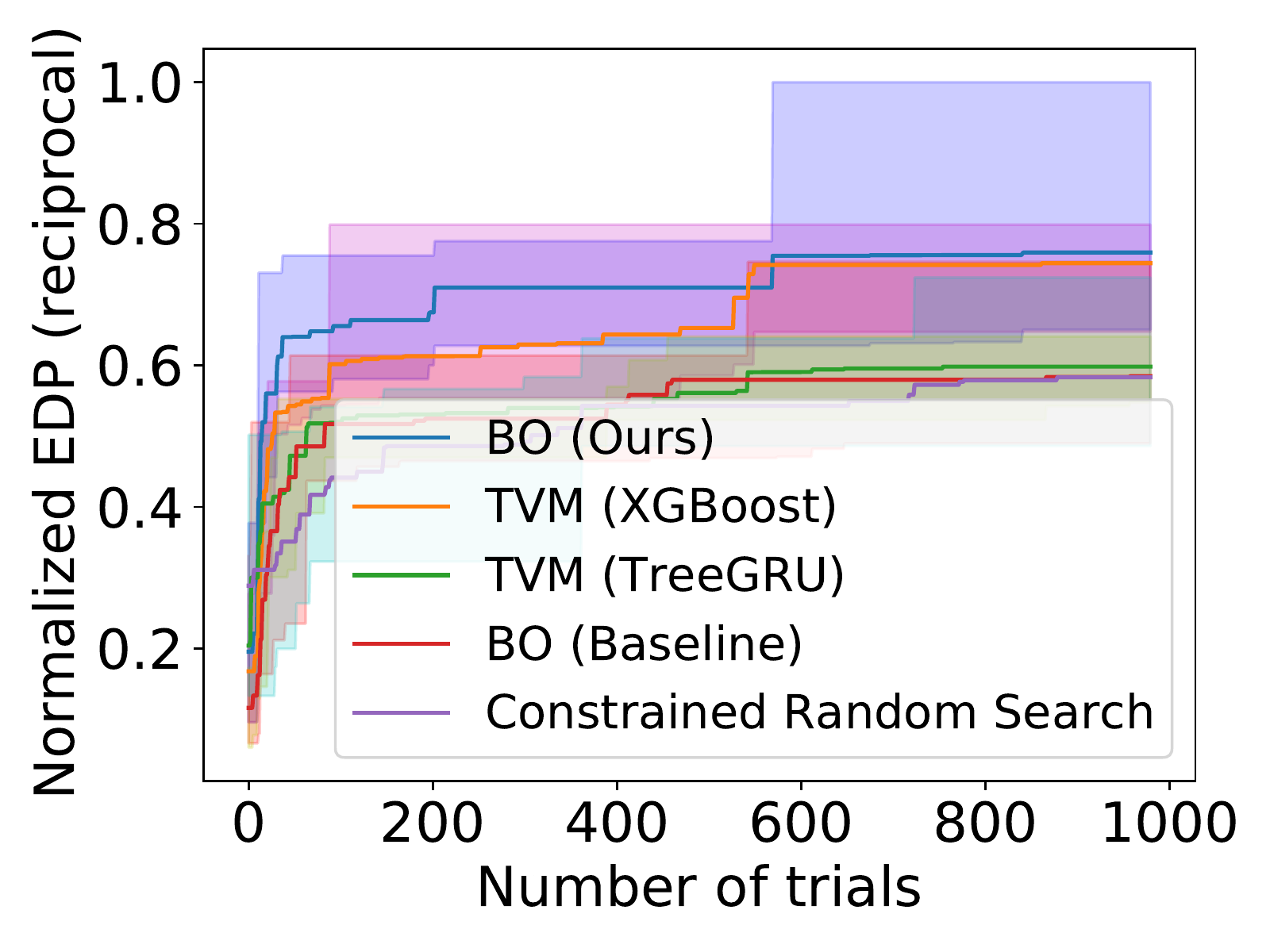}
  \caption{Transformer-K3}
\end{subfigure}
\begin{subfigure}{.3\textwidth}
  \centering
  \includegraphics[width=1\textwidth]{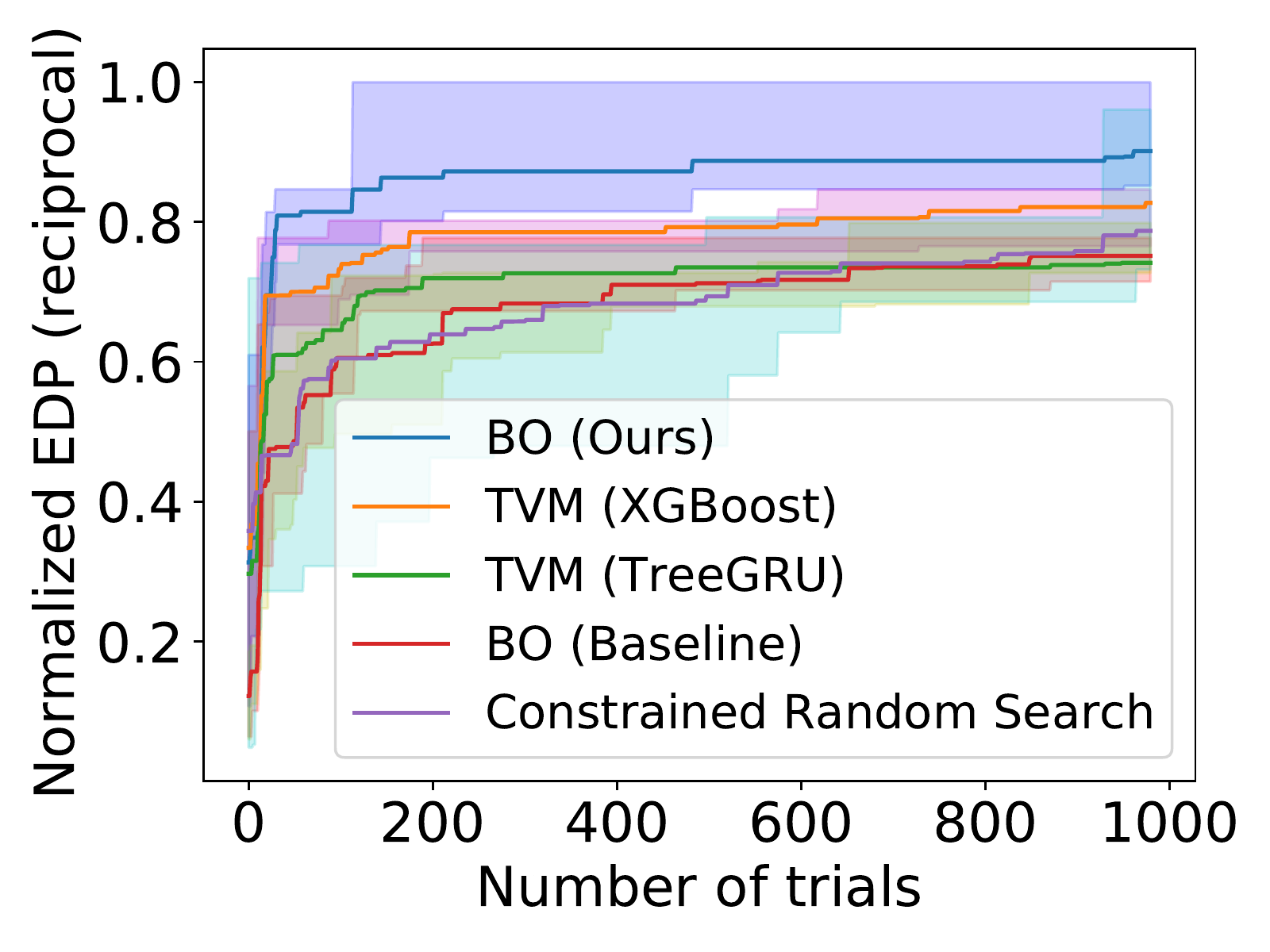}
  \caption{Transformer-K4}
\end{subfigure}
\caption{Software mapping optimization on ResNet, DQN, MLP, and Transformer. The Y-axis shows the reciprocal of energy-delay product (EDP) (normalized against the best EDP value). Higher is better.}
\label{fig:sw_curves_appendix}
\end{figure}

In Figure \ref{fig:sw_curves_appendix} we show more examples of the software optimization over multiple layers of the different architectures. Our Bayesian optimization formulation consistently outperforms the baselines~\citep{chen2018learning}.

\subsection{Ablations}
\begin{figure}[ht]
\centering
\begin{subfigure}{.3\textwidth}
  \centering
  \includegraphics[width=1\textwidth]{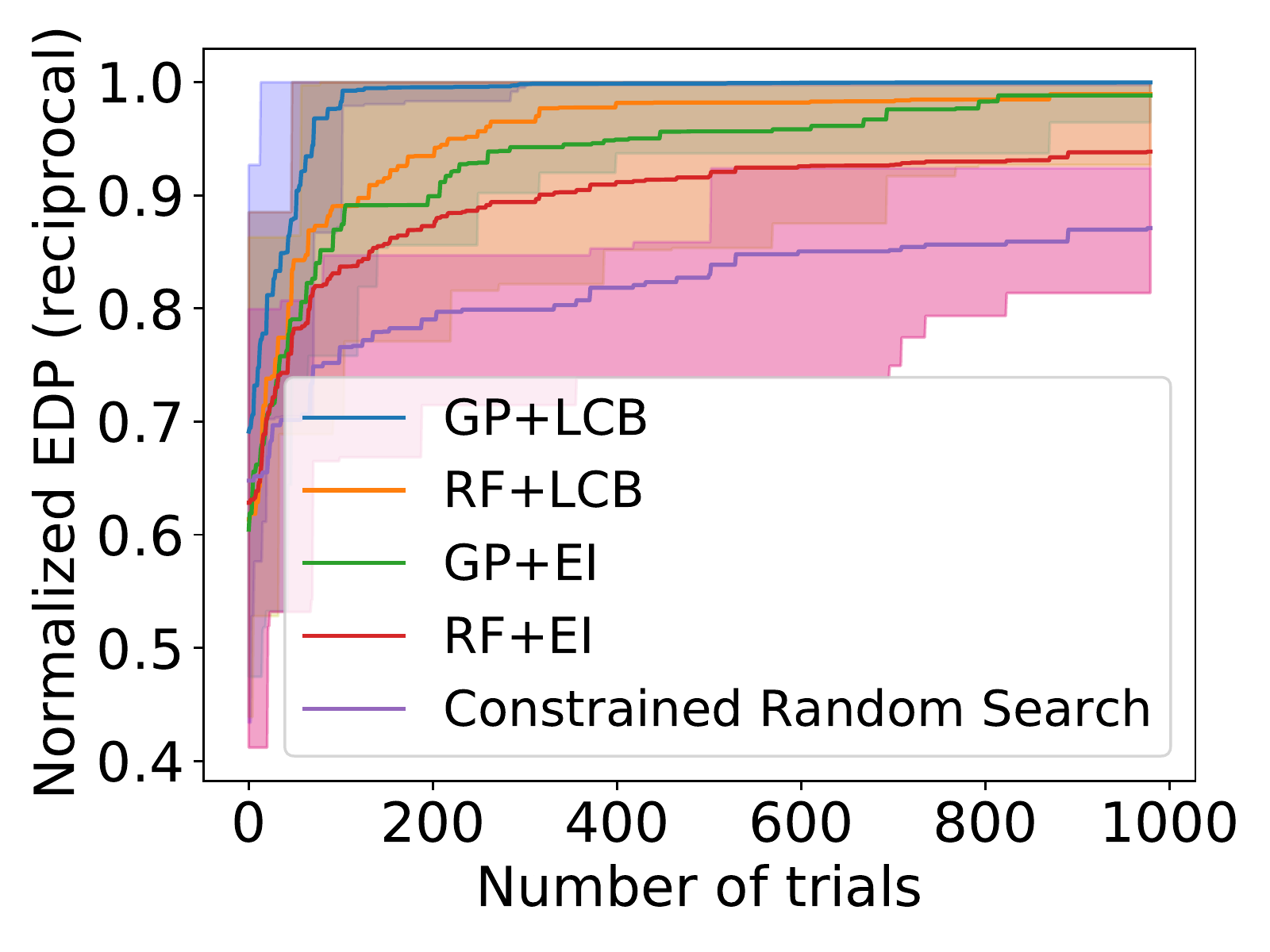}
  \caption{ResNet-K2}
\end{subfigure}
\begin{subfigure}{.3\textwidth}
  \centering
  \includegraphics[width=1\textwidth]{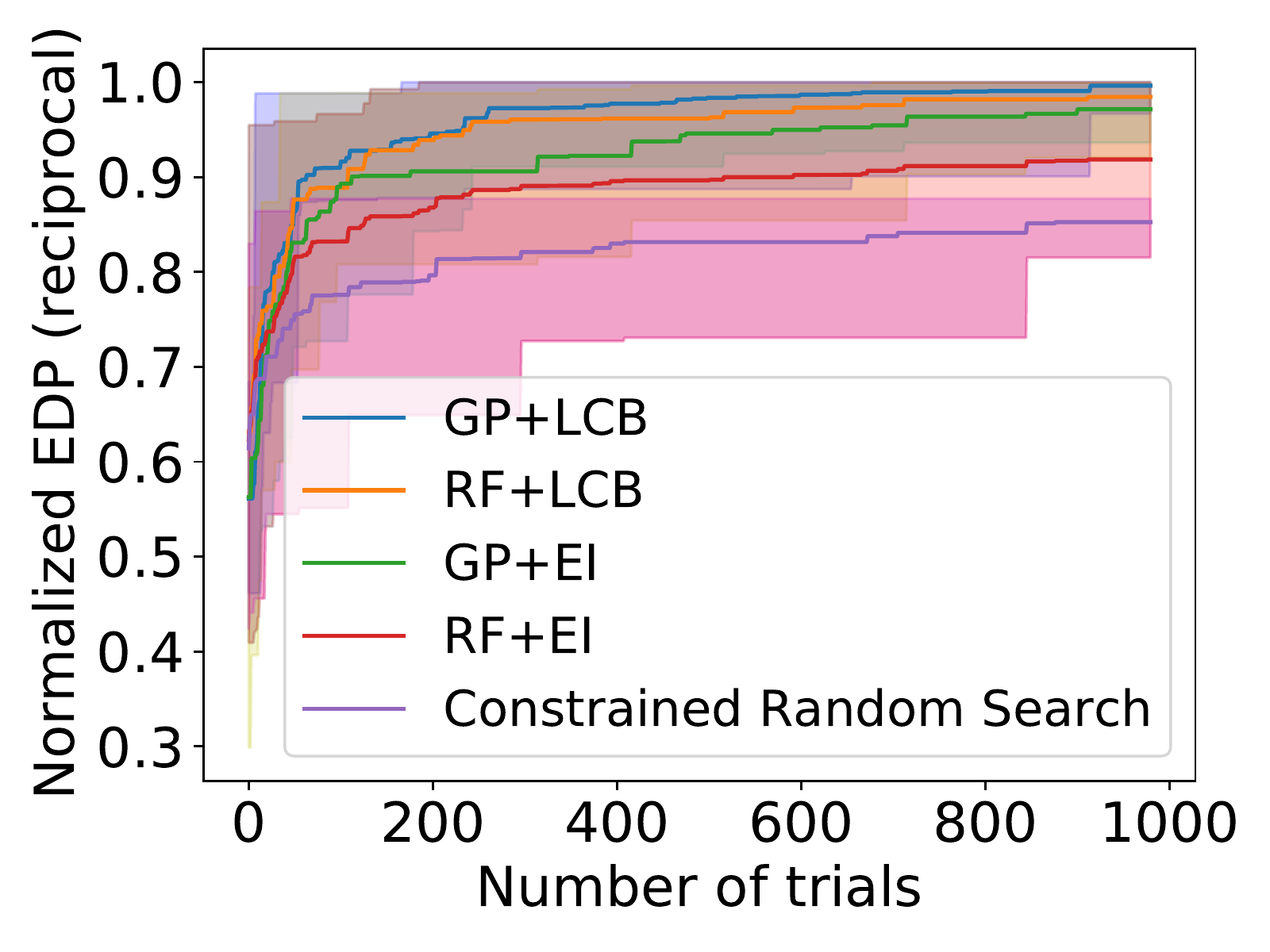}
  \caption{ResNet-K3}
\end{subfigure}
\begin{subfigure}{.3\textwidth}
  \centering
  \includegraphics[width=1\textwidth]{figures/ablation_acquisition/gp-4.pdf}
  \caption{ResNet-K4}
\end{subfigure}
\caption{GP with different surrogate models and acquisition functions.}
\label{fig:bo_ablation_appendix}
\end{figure}

In Figure~\ref{fig:bo_ablation_appendix} we compare different surrogate models and acquisition functions for Bayesian optimization of the software mapping. We found Gaussian processes with LCB to consistently outperform other alternatives.

\begin{figure}[ht]
\centering
\begin{subfigure}{.3\textwidth}
  \centering
  \includegraphics[width=1\textwidth]{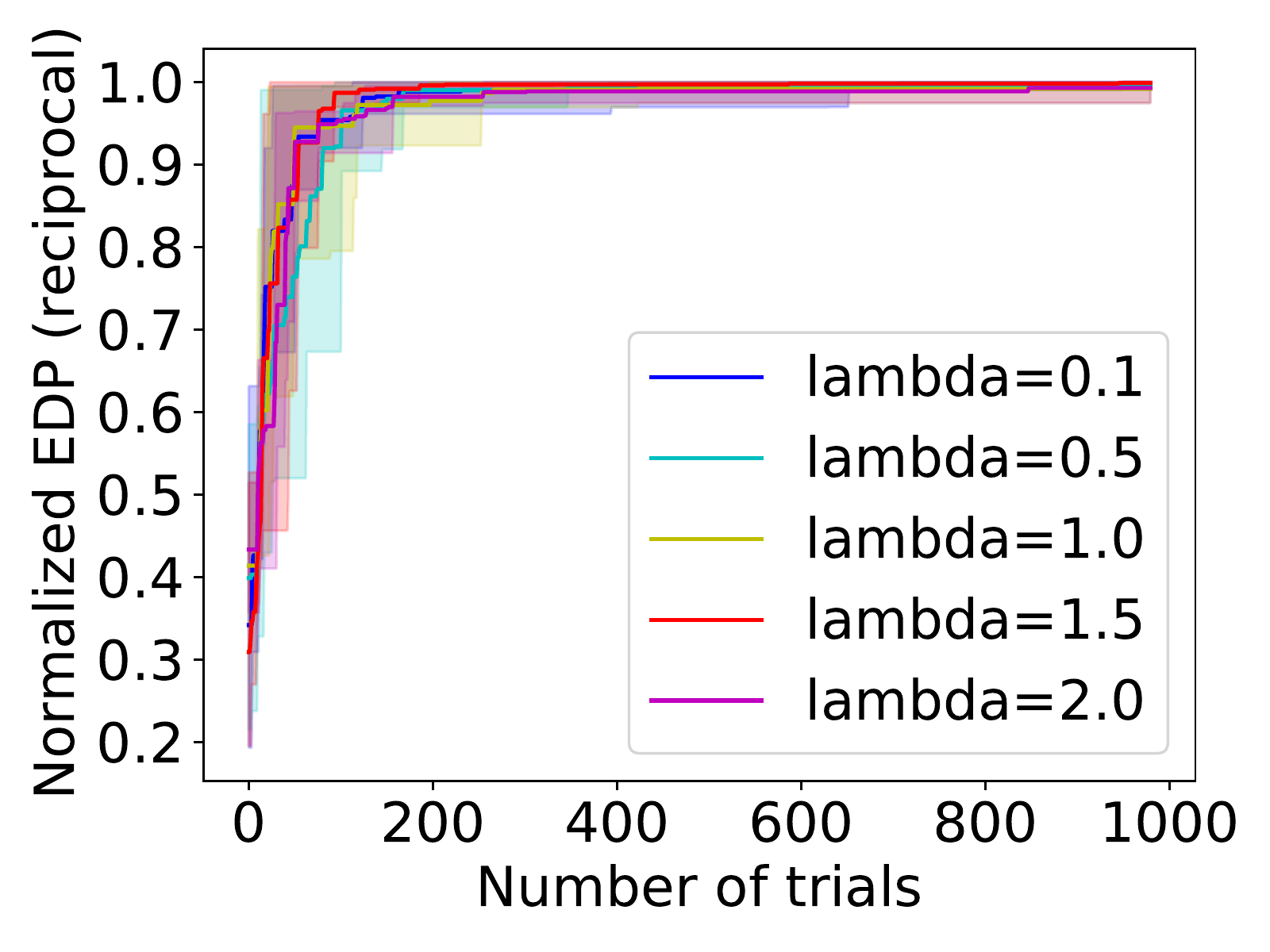}
  \caption{ResNet-K2}
\end{subfigure}
\begin{subfigure}{.3\textwidth}
  \centering
  \includegraphics[width=1\textwidth]{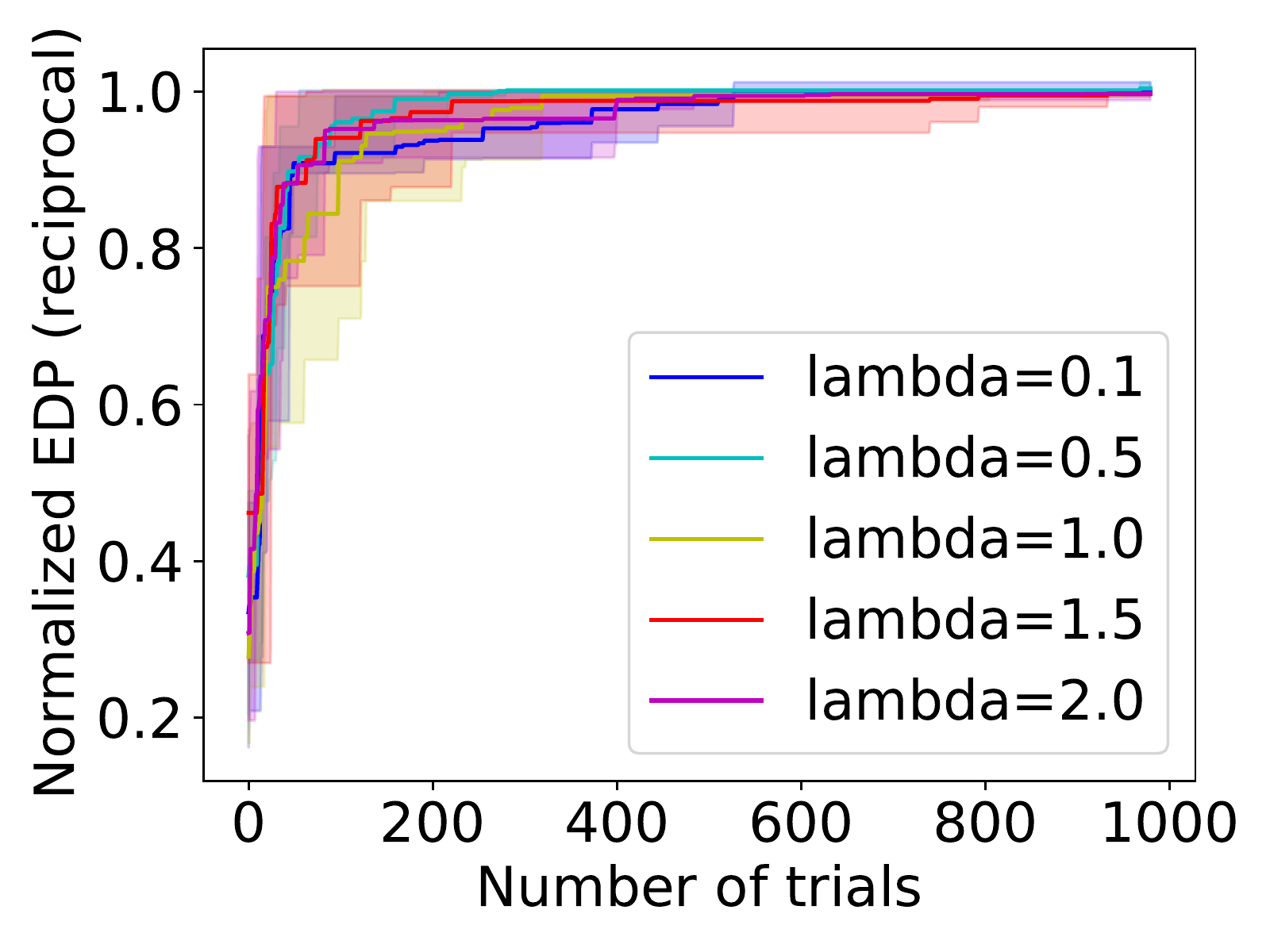}
  \caption{ResNet-K3}
\end{subfigure}
\begin{subfigure}{.3\textwidth}
  \centering
  \includegraphics[width=1\textwidth]{figures/ablation_acquisition/lcb-4.pdf}
  \caption{ResNet-K4}
\end{subfigure}
\caption{LCB acquisition function with different lambda values.}
\label{fig:lcb_ablation_appendix}
\end{figure}

In Figure~\ref{fig:lcb_ablation_appendix} we investigate the robustness of LCB for software optimization using different values of $\lambda$. We found that $\lambda=0.1$ tends to be too greedy, but that above $\lambda=0.5$, LCB tends to be fairly robust.

\end{document}